%% file: acl_latex.tex
\pdfoutput=1

\documentclass[11pt]{article}

\usepackage[final]{acl}
\usepackage{booktabs}
\usepackage{amsmath}
\usepackage{tabularx} 
\usepackage{times}
\usepackage{latexsym}
\usepackage{amssymb} 
\usepackage[T1]{fontenc}
\usepackage[most]{tcolorbox}
\tcbuselibrary{skins, breakable}
\usepackage{float}
\usepackage{multirow}
\usepackage{dblfloatfix}
\usepackage{microtype}
\usepackage{dblfloatfix} 
\usepackage{placeins}
\usepackage{inconsolata}
\usepackage{subcaption} 
\usepackage{url}
\usepackage{caption}
\usepackage{hyperref}
\usepackage{amsmath}
\usepackage{amssymb}
\usepackage{cuted}   
\usepackage{caption} 
\usepackage{mathtools}
\usepackage{amsthm}
\usepackage{amsmath}
\usepackage{xspace}
\usepackage{multirow}
\usepackage{tabularx}
\usepackage{enumitem}
\usepackage[english]{babel}
\usepackage{subcaption}
\usepackage{svg}
\usepackage{caption}
\usepackage{booktabs}
\usepackage{colortbl}
\usepackage{bbding}
\usepackage{graphicx}
\usepackage{amsmath}
\usepackage{algorithm}
\usepackage[noend]{algpseudocode}
\usepackage[most]{tcolorbox}
\usepackage{enumitem}
\usepackage{soul}
\usepackage{cleveref}
\usepackage{wrapfig}
\usepackage{mathtools}
\setlist[itemize]{noitemsep,nolistsep}
\theoremstyle{plain}

\theoremstyle{definition}

\theoremstyle{remark}

\theoremstyle{plain}      
\newtheorem{property}{Property}
\usepackage{pifont}
\usepackage{wasysym}
\newcommand{\method}{\ensuremath{\textnormal{\textsc{RILKE}}}\xspace}

\newcommand{\nop}[1]{}
\usepackage{natbib}

\title{Representation Interventions Enable Lifelong \\ Knowledge Memory Control in LLMs}

\author{Xuyuan Liu\thanks{Work done during an internship at NEC Laboratories America.}$^{1,2}$,
Shengyu Chen$^{2}$,
Xinshuai Dong$^{3}$, 
Yanchi Liu$^{2}$,
Xujiang Zhao$^{2}$, \\
\textbf{
Haoyu Wang$^{2}$,
Yujun Yan$^{1}$,
Haifeng Chen$^{\dag\,,2}$, 
Zhengzhang Chen\thanks{Corresponding authors.}$^{,2}$}\\
$^{1}$Dartmouth College \;
$^{2}$NEC Laboratories America \;
$^{3}$Carnegie Mellon University \\
\texttt{\{xuyuan.liu.gr,yujun.yan\}@dartmouth.edu, zchen@nec-labs.com} 
}

\begin{document}
\maketitle

\input{sections/0_abstract}
\input{sections/1_introduction}

\input{sections/3_preliminaries}

\input{sections/4_method}

\input{sections/5_analysis}

\input{sections/6_res_analysis}

\input{sections/2_related_works}

\input{sections/7_conclusion}
\bibliography{custom}

\appendix
\input{sections/appendix}

\end{document}

%% file: sections/0_abstract.tex
\begin{abstract}

Large language models (LLMs) often produce incorrect or outdated content after being employed. Efficient and accurate knowledge updates without costly retraining are a major challenge. This problem is particularly challenging in \emph{lifelong} settings, where complex, unstructured knowledge must coexist without interference. We introduce \method~(\underline{R}epresentation \underline{I}ntervention for \underline{L}ifelong \underline{K}nowledg\underline{E} Control), a robust and scalable method that treats knowledge control as interventions within the model’s representation space. Leveraging representation-space expressiveness, we identify two key properties enabling \method to achieve fine\mbox{-}grained control over complex, unstructured knowledge while maintaining general utility with frozen base weights. 
During training, \method~learns \emph{paraphrase-robust} and \emph{edit\mbox{-}localized} modules that limit each update to a low-dimensional subspace to minimize cross-edit interference. At inference, a query-adaptive router selects the appropriate module to guide the model’s generation. Across LLaMA and Qwen models, \method scales effectively to large-scale benchmarks, demonstrating high edit success and strong paraphrase generalization while preserving general utility with modest memory overhead. These results show \method~is an effective and scalable solution for lifelong knowledge control in LLMs.

\end{abstract}

%% file: sections/1_introduction.tex
\section{Introduction}

Large language models (LLMs) excel at knowledge-intensive NLP tasks~\citep{madaan2022emnlp,sun-etal-2024-head,chen-etal-2025-large-language}, yet their knowledge is static. Once deployed, they cannot adapt to evolving real-world information, often producing outdated or inaccurate content. While methods like retraining or continual pretraining can update a model, they are computationally prohibitive and prone to catastrophic forgetting~\citep{ke2023iclr, yildiz2025tmlr}. As an alternative, Retrieval-Augmented Generation (RAG) injects new facts at inference time~\citep{chen-etal-2024-lifelong}, but it is susceptible to conflicts with parametric memory~\citep{li2025taming, gutierrez2025rag} and sensitive to retrieval quality~\citep{salemiza2024sigir}. These challenges underscore the need for \textbf{lifelong knowledge memory control}: methods to precisely and efficiently update LLM knowledge with minimal side effects~\citep{thede2025wikibigedit}.

However, existing methods struggle in lifelong edit settings, especially when addressing \emph{unstructured}, \emph{free-form} knowledge that cannot be reduced to simple factual triplets (\textit{i.e.}, Subject--Relation--Object)~\citep{DBLP:conf/iclr/DengWPDSC25}. Parametric approaches, which modify model weights directly~\citep{DBLP:conf/iclr/FangJWMSW0C25},  suffer from \textit{edit collapse}: as new knowledge accumulates, performance degrades until the model becomes unusable~\citep{yang2024acl, nishi2025representation}, a degradation particularly severe when incorporating new unstructured knowledge. Similarly, although effective for accumulation, external-memory methods struggle to capture the nuances of complex information, as their capacity is constrained by learning on the weight space of a single sub-module (\textit{e.g.}, an additional Feedforward network (FFN) layer, as in~\citet{DBLP:conf/nips/0104L0XY0X0C24}).

In contrast to modifying model weights, interventions in the representation space, the hidden states where information is processed, offer a powerful substrate for knowledge control due to their rich semantic structure, yet this space remains largely underexplored~\citep{elhage2021mathematical,DBLP:conf/nips/WuAWGJMP24}. To our knowledge, the only related work on representation-space editing is limited to structured knowledge and supports only single, instantaneous edits~\citep{liu2025unlocking}. It fails to leverage the representation space's potential for complex, unstructured knowledge and cannot handle lifelong settings where edits must accumulate and coexist without interference.

In this work, we unlock the potential of representation interventions for precise, paraphrase-robust, and lifelong knowledge control. We introduce \textbf{RILKE} (\underline{R}epresentation \underline{I}ntervention for \underline{L}ifelong \underline{K}nowledg\underline{E} Control), a framework that controls LLM behavior by intervening in the model's hidden representations. We first identify and validate two key geometric properties of these representations that enable localized and scalable editing. Building on this, we develop a robust training strategy to ensure that edits generalize across paraphrases. For lifelong learning, we design a dynamic routing mechanism that activates the correct intervention at inference time, mitigating interference between edits. Finally, we introduce a shared-subspace intervention that clusters similar edits into a single module, enabling grouped control and significantly improving memory efficiency. Across various models and benchmarks, \method provides a stable, lightweight, and interpretable solution for lifelong knowledge control in LLMs.

We summarize our contributions as follows:
\begin{itemize}[leftmargin=*, itemsep=0pt]
    \item We propose a robust training scheme that achieves precise, generalizable control through LLM representation-space interventions.
    \item We propose a dynamic router that selectively activates relevant interventions while preserving unrelated knowledge and overall utility.
    \item We develop a shared-subspace intervention strategy for memory-efficient, scalable control over large, growing sets of knowledge.
\end{itemize}

%% file: sections/3_preliminaries.tex
\section{Preliminaries}
\label{sec:preliminaries}

This section introduces the notation used throughout the paper, along with the core concepts of LLMs and the Representation Fine-Tuning (ReFT).

Consider a token sequence $\boldsymbol{x}=(x_1,\ldots,x_n)$, where each $x_i$ is an element of a vocabulary $\mathcal{V}$. An LLM  parameterized by $\theta$ defines a joint probability over $\boldsymbol{x}$ through an auto-regressive factorization:

\vspace{-19pt}
\[
\vspace*{-0.3em}
p_\theta(\boldsymbol{x})=\prod_{i=1}^n p_\theta\!\left(x_i \mid \boldsymbol{x}_{<i}\right),\; \boldsymbol{x}_{<i}\!=\!(x_1,\ldots,x_{i-1})
\]

\noindent where $p_\theta\left(x_i \mid \boldsymbol{x}_{<i}\right)$ denotes the model’s predictive distribution over $\mathcal{V}$ for the token at position $i$, conditioned on the prefix $\boldsymbol{x}_{<i}$. For an $L$-layer model, let $\boldsymbol{h}^{l,i}$ be the hidden representation at position $i$ in layer $l$. The next token probability distribution is obtained by applying a softmax function to the linear projection of the hidden state of the final layer $\boldsymbol{h}^{L,i}$, using a weight matrix $\mathbf{W}$:
\vspace{-2pt}
\[
\vspace*{-0.3em}
p_\theta\left(x_i \mid \boldsymbol{x}_{<i}\right)=\operatorname{softmax}\left(\mathbf{W} \boldsymbol{h}^{L,i}\right) .
\]

To generate a sequence $\boldsymbol{x}$, the LLM iteratively computes $p_\theta\left(x_i \mid \boldsymbol{x}_{<i}\right)$,  samples a token $x_i$ at position $i$, and appends it to the context for the next step. This process terminates upon generating a designated end-of-sequence token or when a predefined maximum length is reached.

\paragraph{Representation Fine-Tuning (ReFT)}~\citep{DBLP:conf/nips/WuAWGJMP24} is a fine-tuning approach for LLMs that, instead of updating model weights, learns an \emph{intervention module} to steer intermediate hidden states so they produce the target outputs, leaving the original weights unchanged. Building on the \textit{linear representation hypothesis}~\citep{mikolov-etal-2013-linguistic,10.5555/3692070.3693675} that concepts are encoded in linear subspaces of the hidden space, ReFT specifies a low-dimensional intervention subspace at layer $l$ via $\mathbf{R}^l\in\mathbb{R}^{r\times d} (r\ll d)$, whose rows are constrained to be orthonormal. Given the original layer-$l$ hidden state
$\boldsymbol{h}^{l,i}\in\mathbb{R}^{d}$ for $i$-th token, ReFT first computes a subspace-local shift $(\mathbf{A}^l \boldsymbol{h}^{l,i} + \mathbf{b}^l - \mathbf{R}^l \boldsymbol{h}^{l,i})$, maps it back to the original space with
$\mathbf{R}^{l\top}$, and adds the result to initial representation $\boldsymbol{h}^{l,i}$, yielding
\begin{equation}
   \Phi(\boldsymbol{h}^{l,i}; \phi^l) = \boldsymbol{h}^{l,i}+\mathbf{R}^{l\top}(\mathbf{A}^l \boldsymbol{h}^{l,i}+\mathbf{b}^l-\mathbf{R}^l \boldsymbol{h}^{l,i}),  
   \label{equ:reft}
\end{equation}
where $\phi^l=(\mathbf{R}^l,\mathbf{A}^l,\mathbf{b}^l)$ denotes the learnable parameters of this intervention module. 
Computation in layers beyond $l$ is identical to the base model: let $\mathbf{F}^{>l}$ be the
mapping  from layer $l+1$
to the last layer before the output head, the next-token distribution is formulated as $p_\theta\left(x_i \mid \boldsymbol{x}_{<i}\right)=\operatorname{softmax}\left(\mathbf{W} \mathbf{F}^{>l}(\Phi(\boldsymbol{h}^{l,i}; \phi^l) \right))$.

For brevity, we omit the token and layer indices $(i, l)$ when the context is clear. Following prior work~\citep{DBLP:journals/corr/abs-2310-01405,liu-etal-2025-spectral}, we use the hidden state of the last token to represent the entire sentence (\textit{i.e.}, $\boldsymbol{h}^l_{\boldsymbol{x}} = \boldsymbol{h}^{l,n}$ for $\boldsymbol{x}$ of length $n$).

%% file: sections/4_method.tex
\newlength{\panelheight}
\setlength{\panelheight}{0.30\textwidth}

\begingroup
\setlength{\dbltextfloatsep}{10pt plus 1pt minus 2pt}

\captionsetup[sub]{skip=0pt, belowskip=-5pt}

\begin{figure*}[!t]
  \centering
  \makebox[\textwidth][c]{%
    \subcaptionbox{Distribution of distances between original queries and their paraphrased counterparts,  compared against a baseline of random non-corresponding pairs. Paraphrased queries remain close to the originals in representation space, supporting Prop.~\ref{ass:rephr}.%
    \label{fig:distribution}}{%
      \includegraphics[height=\panelheight,keepaspectratio]{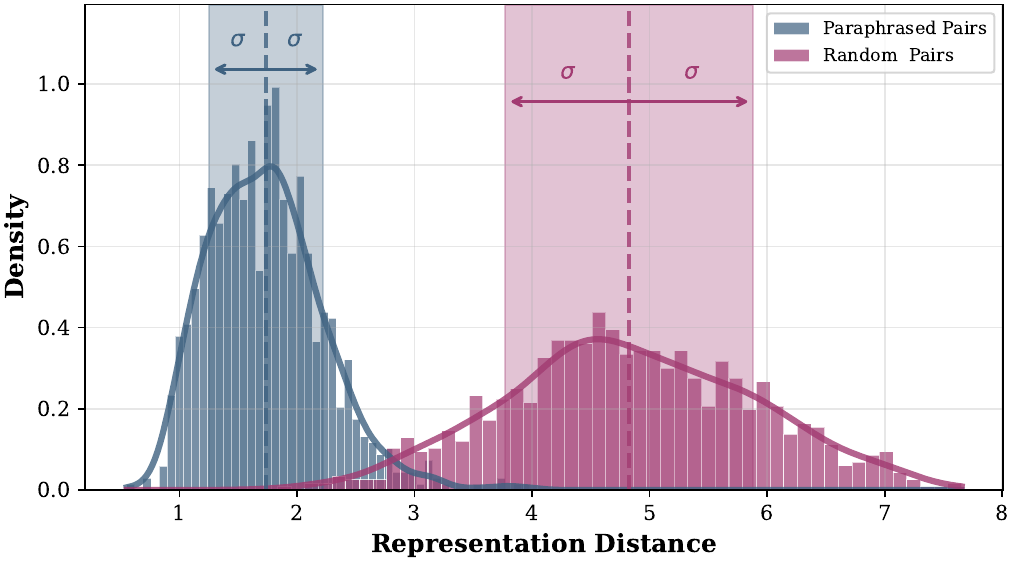}%
    }\hspace{0.8em}%
    \subcaptionbox{Mean \(\mathbf{R}^{l}\)-similarity stratified by representation similarity. Higher representation similarity corresponds to stronger \(\mathbf{R}^{l}\)-alignment, indicating a shared intervention subspace (Prop.~\ref{ass:task}). \label{fig:task_dis}
    }{%
      \includegraphics[height=\panelheight,keepaspectratio]{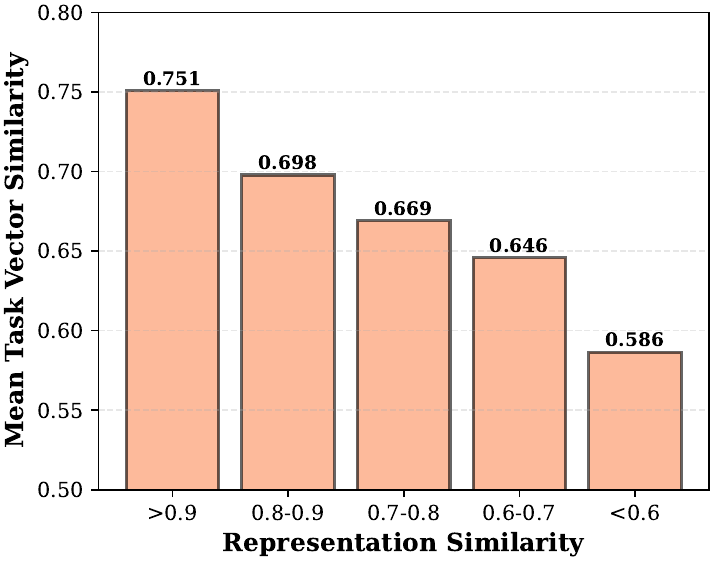}%
    }%
  }
  \vspace{-2pt}
  \caption{\small{Two key properties in LLMs’ representation space. We show that these properties can be utilized to facilitate generalizable, lifelong, and scalable knowledge control in LLMs.}}
  \vspace{-2pt}
\end{figure*}
\endgroup

\section{Motivation: Geometric Properties of LLM Hidden States}
\label{sec:motivation}

LLMs’ hidden states provide a highly expressive mechanism for controlling model behavior~\citep{DBLP:journals/corr/abs-2310-01405,rimsky-etal-2024-steering}, yet existing approaches largely operate at a coarse granularity (\textit{e.g.}, concept-level control)~\citep{DBLP:conf/nips/ArditiOSPPGN24,marks2024the}. Consequently, achieving precise, fine-grained control over individual pieces of knowledge remains challenging, particularly in free-form settings that require lengthy, accurate generation to fully express the knowledge. In such settings, an effective free-form knowledge control method must address three key challenges: \textbf{(i) catastrophic forgetting}, ensuring that newly introduced updates do not interfere with previously learned knowledge; \textbf{(ii) generalizability}, requiring edits to transfer to paraphrased or semantically equivalent queries; and \textbf{(iii) scalability}, supporting a growing number of edits with minimal additional memory overhead.

To address these challenges, we characterize two geometric properties of hidden-state representations. The first, termed \textbf{semantic locality}, captures the tendency for representations to be primarily shaped by semantic content and to remain stable under lexical variation, so long as the underlying semantics are preserved. Formally,
\begin{property}\label{ass:rephr}
Let $\boldsymbol{h}^l_{\boldsymbol{x}}$ denote the layer-$l$ representation of a query $\boldsymbol{x}$. For any paraphrase $\hat{\boldsymbol{x}}$ of $\boldsymbol{x}$ and any semantically unrelated query $\boldsymbol{x}'$, the following holds:
$\|\boldsymbol{h}^l_{\hat{\boldsymbol{x}}}-\boldsymbol{h}^l_{\boldsymbol{x}}\|_2 < \|\boldsymbol{h}^l_{\boldsymbol{x}'}-\boldsymbol{h}^l_{\boldsymbol{x}}\|_2$
\end{property}

This property provides a mechanism to mitigate \textbf{catastrophic forgetting}; by exploiting representation similarity to route queries to disentangled knowledge modules, the system isolates updates and ensures that only the relevant module is activated during inference. Moreover, semantic locality supports \textbf{generalizability}: because paraphrases remain proximal in representation space, an edit applied to the neighborhood of the original query naturally transfers to them. Notably, this behavior is non-trivial—it is driven primarily by semantic similarity rather than token-level overlap, and remains robust even when lexical features are misleading, as shown in App.~\ref{subsec:robust_locality}.

Building on this observation, we further uncover a connection between query semantics and the learned intervention function space. Specifically,

\begin{property}\label{ass:task}
Let $\boldsymbol{x}_i$ and $\boldsymbol{x}_j$ be two semantically related queries. Even when ReFT (Eq.(\ref{equ:reft})) is trained independently for each query, their learned layer-$l$ intervention subspaces remain aligned; that is, the corresponding projection matrices $\mathbf{R}^l_{\boldsymbol{x}_i}$ and $\mathbf{R}^l_{\boldsymbol{x}_j}$ are highly similar.
\end{property}

In other words, semantically related knowledge tends to concentrate within a shared low-dimensional subspace. This structure allows a single module to identify such a subspace and efficiently store and apply multiple semantically similar edits within it, thereby improving \textbf{scalability} without increasing interference across edits.

We empirically validate these properties. Fig.~\ref{fig:distribution} corroborates Prop.~\ref{ass:rephr} by showing a clear separation between the $\ell_2$-distance distributions of paraphrased query pairs and those of randomly paired queries; moreover, this separation remains robust under substantial lexical variation. For Prop.~\ref{ass:task}, we analyze the relationship between representation similarity and the alignment of learned intervention subspaces. Specifically, we compute the cosine similarity between subspace projection matrices $\mathbf{R}^{l}$ (see Eq.(\ref{equ:reft})) within groups stratified by representation similarity, and observe that higher representation similarity consistently corresponds to stronger subspace alignment (Fig.~\ref{fig:task_dis}). These results support the feasibility of controlling semantically related knowledge through grouped interventions.

\begin{figure*}[htp]

    \centering
    \includegraphics[width=0.89\linewidth]{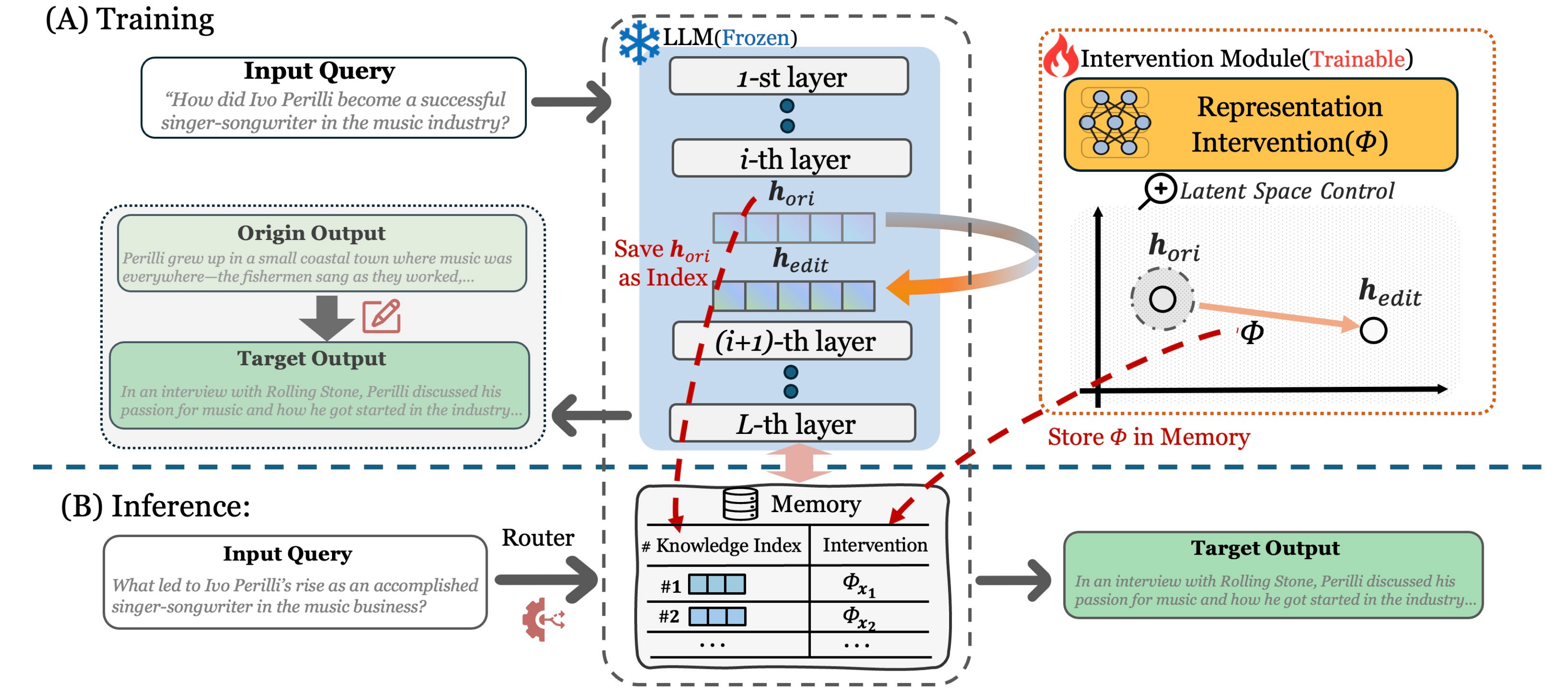}
    \caption{\small {Overview of the \textsc{RILKE} framework. During training, the intervention module $\Phi$ maps $\boldsymbol{h}_{\text{ori}}$ to the target $\boldsymbol{h}_{\text{edit}}$, with $\boldsymbol{h}_{\text{ori}}$ stored as the knowledge index. During inference, the router selects the intervention module $\Phi$ whose index is closest to the input query’s representation to perform a targeted edit, enabling the model to generate the desired output.}}
    \label{fig:framwork}
    \vspace{-3pt}
\end{figure*}

\section{Method: Towards Generalizable, Scalable, Lifelong Knowledge Control}
\nop{Motivated by the observations in Sec.~\ref{sec:motivation}, we propose \textbf{RILKE} (\textbf{R}epresentation \textbf{I}ntervention for \textbf{L}ifelong \textbf{K}nowl\textbf{E}dge Control), a framework for lifelong control of LLMs. During training, \textsc{RILKE} assigns a dedicated representation-intervention module to each edited knowledge item and stores updates in a distributed manner, avoiding catastrophic forgetting and cross-edit interference. In inference, a query-adaptive router activates the most relevant module, steering generation to the target output while preserving unrelated knowledge. We further introduce a batched training scheme that clusters semantically related knowledge and trains shared intervention modules, improving memory efficiency and enabling scalable knowledge management within \textsc{RILKE}.}

To address the challenges of \textit{catastrophic forgetting}, \textit{generalizability}, and \textit{scalability} in knowledge editing, we introduce \textsc{RILKE} (\textbf{R}epresentation \textbf{I}ntervention for \textbf{L}ifelong \textbf{K}nowledg\textbf{E} Control). \textsc{RILKE} achieves lifelong knowledge control through a modular approach:  during training, \textsc{RILKE} mitigates forgetting by isolating updates within dedicated intervention modules. At inference, a query-adaptive router ensures precise generation by activating the appropriate module only when relevant. Finally, to scale this approach, we incorporate a batched training scheme that clusters semantically similar edits into shared modules, thereby improving memory efficiency.

\subsection{Consistency-Robust Training for Generalizable Knowledge Control}

\label{subsec:single_edit}

Knowledge-editing methods always struggle with \emph{generalizability}: while an edit may successfully update a specific query, it often fails on paraphrases of that same query.  Consider an unstructured knowledge pair \((\boldsymbol{x},\boldsymbol{y})\) comprising an input query \(\boldsymbol{x}\) and a target response \(\boldsymbol{y}\). 
Vanilla ReFT exemplifies this issue,  as it learns an intervention $\phi^l_{\boldsymbol{x}}$ strictly conditioned on the representation $\boldsymbol{h}^{l}_{\boldsymbol{x}}$ of the original input. Consequently, a semantically equivalent paraphrase $\hat{\boldsymbol{x}}$, which yields a proximal yet distinct representation $\boldsymbol{h}^{l}_{\hat{\boldsymbol{x}}}$, may fail to trigger the edit.

Building on Prop.~\ref{ass:rephr} and Eq. (\ref{equ:reft}), we assume the layer-$l$ representation of a paraphrased query $\hat{\boldsymbol{x}}$ resides within an $\varepsilon$-ball centered at $\boldsymbol{h}^{l}_{\boldsymbol{x}}$, \textit{i.e.},
\(
 \| \boldsymbol{h}^{l}_{\hat{\boldsymbol{x}}}-\boldsymbol{h}^{l}_{\boldsymbol{x}} \|_2 \le \varepsilon .
\)
We therefore impose a consistent intervention target throughout this region. Applying a first-order expansion to the ReFT map with respect to the deviation $\varepsilon$ yields
\resizebox{0.49\textwidth}{!}{$\displaystyle
\Phi(\boldsymbol{h}^{l}_{\boldsymbol{x}}+\varepsilon;\phi^{l}_{\boldsymbol{x}})
=\Phi(\boldsymbol{h}^{l}_{\boldsymbol{x}};\phi^{l}_{\boldsymbol{x}})
+{\big(\mathbf{I}+\mathbf{R}_{\boldsymbol{x}}^{l\top}(\mathbf{A}_{\boldsymbol{x}}^l-\mathbf{R}_{\boldsymbol{x}}^l)\big)}\varepsilon
$}
\noindent
implying that the variation induced by the paraphrase is governed by \(\big(\mathbf{I}+\mathbf{R}_{\boldsymbol{x}}^{l\top}(\mathbf{A}_{\boldsymbol{x}}^l-\mathbf{R}_{\boldsymbol{x}}^l)\big)\varepsilon\).
To mitigate this effect, we enforce consistency of the final vocabulary distribution within the ball. We can see the edited predictive distribution for initial representation \(\boldsymbol{h}^{l}_{\boldsymbol{x}}\) is:
\vspace{-4pt}
\[
p_{\theta,\phi^{l}_{\boldsymbol{x}}}(\cdot\mid \boldsymbol{x})
=\operatorname{softmax}\!\big(\mathbf{W}\mathbf{F}^{>l}(\,\Phi(\boldsymbol{h}^{l}_{\boldsymbol{x}};\phi^{l}_{\boldsymbol{x}}))\big).
\]
\noindent
Then, we draw \(\varepsilon\sim\mathcal{Q}\) (\textit{e.g.}, \(\mathcal{N}(0,\sigma^2\mathbf{I})\)) on \(\boldsymbol{h}^{l}_{\boldsymbol{x}}\) and form a perturbed branch:
\vspace{-4pt}
\[
p_{\theta,\phi^{l}_{\boldsymbol{x}}}^{(\varepsilon)}(\cdot\mid \boldsymbol{x})
=\operatorname{softmax}\!\big(\mathbf{W}\mathbf{F}^{>l}(\,\Phi(\boldsymbol{h}^{l}_{\boldsymbol{x}}+\varepsilon;\phi^{l}_{\boldsymbol{x}}))\big).
\]
We penalize the discrepancy between the two distributions using the KL divergence:
\vspace{-4pt}
\[
\mathcal{L}_{\mathrm{robu}}(\boldsymbol{x};\phi^{l}_{\boldsymbol{x}})
=\mathrm{KL}\!\left(p_{\theta,\phi^{l}_{\boldsymbol{x}}}(\cdot\mid \boldsymbol{x})\big\|p_{\theta,\phi^{l}_{\boldsymbol{x}}}^{(\varepsilon)}(\cdot\mid \boldsymbol{x})\right).
\]
\noindent We incorporate this robustness term into the standard language modeling objective, which minimizes cross-entropy loss in a teacher-forcing manner, as in vanilla ReFT. The general objective for updating the knowledge item $(\boldsymbol{x},\boldsymbol{y})$ with a single-knowledge intervention module is, therefore,
\vspace{-0.8em}
\begin{equation}
\vspace*{-0.05em} 
\begin{aligned}
\mathcal{L}(\phi^{l}_{\boldsymbol{x}})
&=-
\sum_{i=1}^{|\boldsymbol{y}|} 
   \log p_{\theta,\phi^{l}_{\boldsymbol{x}}}\!\left(y_i \mid \boldsymbol{x}, \boldsymbol{y}_{<i}\right)
\\
& \hspace{-3pt}
+\lambda_{\mathrm{robu}}\,
\mathbb{E}_{\varepsilon\sim\mathcal{Q}}\!\big[\mathcal{L}_{\mathrm{robu}}(\boldsymbol{x};\phi^{l}_{\boldsymbol{x}})
\big],
\end{aligned}
\label{equ:robust}
\end{equation}

\noindent
where regularization promotes the generalization of edited knowledge to paraphrased queries.

\subsection{Query-Adaptive Routing for Lifelong Knowledge Control}
\label{subsec:single}

To mitigate catastrophic forgetting, we freeze the base model and train a dedicated intervention module $\phi^l_{\boldsymbol{x}}$ using the method in Sec.~\ref{subsec:single_edit} for each knowledge instance $\boldsymbol{x}$.  While this effectively isolates edits, it introduces a new challenge at inference: the model must select and then apply the appropriate intervention for a given prompt. To address this, we leverage Prop.~\ref{ass:rephr} to design a router that directs incoming queries to their corresponding modules.           

Specifically, we construct a routing index  $\{\boldsymbol{h}^{l}_{\boldsymbol{x}_j}\}_{j=1}^m$ using the layer-$l$ representations of all $m$ training examples in the edit dataset.
Since the base model is frozen and interventions function only beyond layer $l$, these representations will not be affected by the training of interventions and act as a stable key space. Each index is then linked to its specific trained intervention \(\{\phi^{l}_{\boldsymbol{x}_j}\}_{j=1}^m\). At inference, given a query \(\hat{\boldsymbol{x}}\) with key \(\boldsymbol{h}^{l}_{\hat{\boldsymbol{x}}}\), the router $\pi(\cdot)$ identifies the stored representation $\boldsymbol{x}_j$ that maximizes the cosine similarity with the query:
\vspace{-3pt}
\[
\vspace*{-0.2em} 
\pi(\boldsymbol{h}^{l}_{\hat{\boldsymbol{x}}})
=\operatorname*{arg\,max}_{\boldsymbol{x}_j,\, 1\le j\le m}\;
\frac{\left\langle \boldsymbol{h}^{l}_{\hat{\boldsymbol{x}}},\, \boldsymbol{h}^{l}_{\boldsymbol{x}_j}\right\rangle}
{\|\boldsymbol{h}^{l}_{\hat{\boldsymbol{x}}}\|_2\;\|\boldsymbol{h}^{l}_{\boldsymbol{x}_j}\|_2}.
\]
Subsequently, \textsc{RILKE} applies the intervention $\Phi(\boldsymbol{h}^{l}_{\hat{\boldsymbol{x}}}; \phi^{l}_{\pi(\boldsymbol{h}^{l}_{\hat{\boldsymbol{x}}})})$ to enact the targeted edit, provided that the maximum similarity score exceeds the predefined relevance threshold $\tau_{\mathrm{sim}}$. If this condition is not met, no intervention is performed.

Importantly, with this strategy, each training query is deterministically assigned to its corresponding intervention module, as its key exactly matches the targeted index. For unseen queries (\textit{e.g.}, paraphrased queries), we observe that most are still routed to the correct module even after large-scale editing, further supporting Prop.~\ref{ass:rephr}.

\subsection{Shared Subspace Intervention for Memory-efficient Management}
\label{subsec:shared}

Lifelong settings require scalable editing of massive knowledge bases. While assigning a dedicated intervention module to each knowledge instance affords fine-grained control, this approach scales poorly, incurring a memory cost that grows linearly with the number of edits and creating a significant bottleneck during training. To address this scalability challenge, we build on the insight from Prop.~\ref{ass:task}, which establishes that semantically related edits can share a common intervention subspace. We therefore propose a memory-efficient alternative: cluster similar knowledge instances and train a single shared intervention module for each cluster.

We partition knowledge instances into semantically homogeneous, size-bounded groups to ensure that a single intervention subspace can serve each group effectively. Concretely, we impose two constraints on every cluster: (i) a within-cluster similarity lower bound $\tau_{\mathrm{sim}}\in(0,1)$ (or equivalently, a merge threshold $d_{\mathrm{thr}}=1-\tau_{\mathrm{sim}}$), and (ii) a maximum cluster size $s_{\max}$.  Let $m$ denote the number of knowledge items and collect their layer-$l$ representations into $\mathbf{H}=\big[\boldsymbol{h}^{l}_{\boldsymbol{x}_1}\,\,\boldsymbol{h}^{l}_{\boldsymbol{x}_2}\,\,\cdots\,\,\boldsymbol{h}^{l}_{\boldsymbol{x}_m}\big]$. We run hierarchical agglomerative clustering (HAC) over the columns of $\mathbf{H}$ using an initial similarity lower bound $\tau_{\min}$ to obtain provisional clusters. Any cluster exceeding $s_{\max}$ is then recursively refined by increasing the similarity floor and re-running HAC within that cluster.  The procedure terminates when all clusters satisfy $|\mathcal{C}_c|\le s_{\max}$, yielding $k$ clusters $\{\mathcal{C}_c\}_{c=1}^k$ that are both semantically coherent and size\mbox{-}controlled (see details in Algo.~\ref{alg:cluster}). 

Given the clustered knowledge, we train, for each cluster $\mathcal{C}_i \in \{\mathcal{C}_c\}$,~
a cluster-shared intervention module using all $\boldsymbol{x}_j \in \mathcal{C}_i$ under the objective in Sec.~\ref{subsec:single_edit}, yielding
$\phi^{l}_{\mathcal{C}_i}=\big(\mathbf{R}^{l}_{\mathcal{C}_i},\,\mathbf{A}^{l}_{\mathcal{C}_i},\,\mathbf{b}^{l}_{\mathcal{C}_i}\big).$
Let $\kappa(\boldsymbol{x}_j)$ be the mapping from 
$\boldsymbol{x}_j$ to its corresponding cluster $\mathcal{C}_i$ (\textit{i.e.}, $\kappa : \{\boldsymbol{x}\} \to \{\mathcal{C}_1, \dots, \mathcal{C}_k\}, \;\kappa(\boldsymbol{x_j})=\mathcal{C}_i$ such that $\boldsymbol{x_j}\in \mathcal{C}_i$).
Then, at inference, for query $\hat{\boldsymbol{x}}$ with key $\boldsymbol{h}^{l}_{\hat{\boldsymbol{x}}}$, router $\hat{\pi}(\cdot)$ maps it to the corresponding intervention by identifying the cluster to which the closest knowledge item belongs:
\vspace{-2pt}
\[
\vspace*{-0.3em}
\begin{aligned}
\hat{\pi}(\boldsymbol{h}^{l}_{\hat{\boldsymbol{x}}}) \;=\;
\kappa\!\left(
\operatorname*{arg\,max}_{\boldsymbol{x}_j,\, 1\le j\le m}
\frac{\left\langle \boldsymbol{h}^{l}_{\hat{\boldsymbol{x}}},\, \boldsymbol{h}^{l}_{\boldsymbol{x}_j}\right\rangle}
{\|\boldsymbol{h}^{l}_{\hat{\boldsymbol{x}}}\|_2\;||\boldsymbol{h}^{l}_{\boldsymbol{x}_j}\|_2}
\right).
\end{aligned}
\]
\noindent
Finally, the corresponding cluster intervention 
\(\Phi(\boldsymbol{h}^{l}_{\hat{\boldsymbol{x}}};\phi^{l}_{{\hat{\pi}(\boldsymbol{h}^{l}_{\hat{\boldsymbol{x}}}) }})\) is applied
for the target edit.

Taken together, these strategies yield \textsc{RILKE}, which unifies (i) robust representation interventions for precise, paraphrase-generalizable knowledge control; (ii) an adaptive inference-time router that activates the appropriate module; and (iii) cluster-level interventions that manage semantically similar knowledge for memory-efficient scalability, thereby collectively enabling lifelong control of unstructured knowledge in LLMs.

%% file: sections/5_analysis.tex
\section{Experiment}

We evaluate the \textsc{RILKE} framework on public benchmarks for knowledge control, focusing on the following research questions:

\begin{itemize}[leftmargin=*,label={}]
    \item \textbf{RQ1:} How does \textsc{RILKE} compare to prior methods in lifelong knowledge learning?
    \item \textbf{RQ2:} How does \textsc{RILKE} leverage shared-space interventions for efficient knowledge control?      
    \item \textbf{RQ3:} How does \textsc{RILKE} facilitate precise and generalizable knowledge control?

\end{itemize}

\subsection{Experimental Setup}

\paragraph{LLMs \& Baselines.}
We evaluate the \textsc{RILKE} framework on popular off-the-shelf LLMs: \texttt{Llama-3.1-8B-Instruct}~\citep{grattafiori2024llama} and \texttt{Qwen2.5-7B-Instruct}~\citep{Yang2024Qwen25TR}. We compare our approach against a diverse set of model editing baselines, including locate-then-edit methods (MEMIT~\citep{DBLP:conf/nips/MengBAB22}, UnKE~\citep{DBLP:conf/iclr/DengWPDSC25}, AnyEdit~\citep{DBLP:journals/corr/abs-2502-05628}) and memory-based approaches (FT-L~\citep{zhu2020modifying}, GRACE~\citep{NEURIPS2023_95b6e2ff}, WISE~\citep{DBLP:conf/nips/0104L0XY0X0C24}) (details in App.~\ref{subsec:baseline}). To ensure fairness, we utilize the standard configurations and implementations from EasyEdit\footnote{\url{https://github.com/zjunlp/EasyEdit}} for all compared methods.

\paragraph{Datasets \& Metrics.}
To assess knowledge editing efficacy, we utilize the UnKE~\citep{DBLP:conf/iclr/DengWPDSC25} and EditEverything~\citep{DBLP:journals/corr/abs-2502-05628} datasets. Following established protocols from prior work \citep{DBLP:journals/corr/abs-2502-05628}, we set the temperature to 0.001 for deterministic generation. We report:
(i) \emph{Lexical Similarity} (Rouge-L), which captures the n-gram overlap between the generated response and the reference; and
(ii) \emph{Semantic Similarity} (BertScore), measured via the cosine similarity of sentence embeddings\footnote[2]{We use \href{https://huggingface.co/sentence-transformers/all-MiniLM-L6-v2}{\texttt{all-MiniLM-L6-v2}} to align with prior work. } to evaluate alignment at the semantic level. We assess both \textit{edit efficacy} (performance on original training queries) and \textit{generalization} (performance on unseen paraphrased queries). Furthermore, to assess the preservation of general capabilities and ensure edit locality, we compare performance on MMLU~\citep{DBLP:conf/iclr/HendrycksBBZMSS21} pre- and post-edit.  The difference serves as a metric for the impact of the edit on unrelated knowledge, with a smaller difference indicating that the editing is localized and does not harm the general utility of the original model. To further demonstrate scalability and broader feasibility, we extend our evaluation to the ZsRE dataset~\citep{DBLP:conf/conll/LevySCZ17} with 3,000 edits to benchmark performance in a structured editing setting.  All metrics are averaged over the evaluation set.

\subsection{\textbf{RQ1}:  \textsc{RILKE}  Enables Lifelong Control }
\label{subsec:rq1}
Consistent with prior work~\citep{DBLP:conf/iclr/DengWPDSC25,DBLP:journals/corr/abs-2502-05628}, we evaluate lifelong knowledge control using a sequential protocol with a batch size of 1 to simulate continuous updates. To mitigate spurious activations, we set the gating similarity threshold $\tau_{\mathrm{sim}}$ to $0.9$, below which the query is considered irrelevant knowledge. For unstructured knowledge, we report edit efficacy and generalization at steps 10, 100, and 1,000, together with MMLU accuracy after 1,000 edits to assess the retention of general capabilities following extensive editing. We also evaluate on ZsRE dataset with 3,000 edits. All results are shown in Tab.~\ref{tab:main_result}.

\begin{table*}[!t]
\vspace{-3pt}
\caption{Results on the UnKE dataset. $T$ denotes the number of sequential edits. \emph{Ori.} reports performance on the original (edited) queries; \emph{Para.} reports generalization to paraphrased queries; and \emph{Util.} reports accuracy on MMLU. Additional results on the ZsRE dataset report reliability (\emph{Rel.}), generalization (\emph{Gen.}), and locality (\emph{Loc.}).}

\centering

\setlength{\tabcolsep}{4.4pt}
\renewcommand{\arraystretch}{1.25}
\setlength{\aboverulesep}{0.2ex}
\setlength{\belowrulesep}{0.2ex}
\setlength{\cmidrulekern}{1pt}

\resizebox{0.99\linewidth}{!}{
\begin{tabular}{@{}l|cccc|cccc|ccccc|cccc@{}} 
\toprule
\multirow{3}{*}{\textbf{Method}}
    & \multicolumn{4}{c|}{$T=10$}
    & \multicolumn{4}{c|}{$T=100$}
    & \multicolumn{5}{c|}{$T=1{,}000$}
    & \multicolumn{4}{c}{ZsRE($T=3{,}000$)} \\
\cmidrule(lr){2-5}\cmidrule(lr){6-9}\cmidrule(lr){10-14}\cmidrule(lr){15-18}
    & \multicolumn{2}{c}{\textbf{Ori.}} & \multicolumn{2}{c|}{\textbf{Para.}}
    & \multicolumn{2}{c}{\textbf{Ori.}} & \multicolumn{2}{c|}{\textbf{Para.}}
    & \multicolumn{2}{c}{\textbf{Ori.}} & \multicolumn{2}{c|}{\textbf{Para.}} & \textbf{Util.}
    & \multirow{2}{*}{\textbf{Rel.}} & \multirow{2}{*}{\textbf{Gen.}} & \multirow{2}{*}{\textbf{Loc.}} & \multirow{2}{*}{\textbf{Avg.}} \\
\cmidrule(lr){2-3}\cmidrule(lr){4-5}\cmidrule(lr){6-7}\cmidrule(lr){8-9}
\cmidrule(lr){10-11}\cmidrule(lr){12-13}\cmidrule(lr){14-14}

& \textbf{BertS} & \textbf{RougeL} & \textbf{BertS} & \textbf{RougeL}
& \textbf{BertS} & \textbf{RougeL} & \textbf{BertS} & \textbf{RougeL}
& \textbf{BertS} & \textbf{RougeL} & \textbf{BertS} & \textbf{RougeL} & \textbf{MMLU}
& & & & \\
\midrule

\multicolumn{10}{l}{\textit{Based on \textsc{LLaMA3.1-8B-Instruct}}} & & & & 0.633 \\

\midrule

\textbf{FT-L}
& 0.059 & 0.010 & 0.005 & 0.070
& 0.120 & 0.023 & 0.126 & 0.020
& 0.112 & 0.030 & 0.111 & 0.031 & 0.226
& 0.05 & 0.01 & 0.04 & 0.03 \\

\textbf{MEMIT}
& 0.754 & 0.558 & \underline{0.720} & 0.571
& 0.195 & 0.151 & 0.178 & 0.153
& 0.033 & 0.145 & 0.034 & 0.142 & 0.188
& 0.00 & 0.00 & 0.00 & 0.00 \\

\textbf{GRACE}
& \underline{0.886} & \underline{0.878} & 0.650 & 0.201
& \underline{0.909} & \underline{0.786} & 0.605 & 0.172
& \underline{0.810} & \underline{0.763} & 0.521 & 0.144  & \underline{0.594}
& 0.46 & 0.01 & \textbf{1.00} & 0.49 \\

\textbf{UnKE}
& 0.627 & 0.442 & 0.599 & 0.373
& 0.250 & 0.202 & 0.294 & 0.210
& 0.013 & 0.080 & 0.017 & 0.070 & 0.126
& 0.02 & 0.02 & 0.01 & 0.02 \\

\textbf{AnyEdit}
& 0.359 & 0.237 & 0.355 & 0.233
& 0.066 & 0.095 & 0.049 & 0.097
& 0.012 & 0.164 & 0.005 & 0.161 & 0.218
& 0.01 & 0.01 & 0.00 & 0.01 \\

\textbf{WISE}
& 0.669 & {0.636} & {0.660} & \underline{0.614}
& 0.672 & 0.664 & \underline{0.669} & \underline{0.598}
& 0.681 & {0.661} & \underline{0.673} & \underline{0.623} & {0.584}
& \underline{0.62} & \underline{0.60} & \textbf{1.00}& 0.73 \\

\midrule

\textbf{RILKE}
& \textbf{1.000} & \textbf{1.000} & \textbf{0.998} & \textbf{0.990}
& \textbf{1.000} & \textbf{1.000} & \textbf{0.984} & \textbf{0.942}
& \textbf{1.000} & \textbf{1.000} & \textbf{0.963} & \textbf{0.882} & \textbf{0.622}
& \textbf{0.99} & \textbf{0.71} & \underline{0.94} & \textbf{0.88} \\

\midrule

\multicolumn{10}{l}{\textit{Based on \textsc{Qwen2.5-7B-Instruct}}} & & & & 0.713 \\

\midrule

\textbf{UnKE}
& 0.825 & 0.430 & {0.777} & 0.405
& {0.653} & {0.375} & {0.640} & {0.382}
& {0.039} & 0.073 & {0.033} & 0.066 & 0.130
& 0.01 & 0.01 & 0.00 & 0.01 \\

\textbf{AnyEdit}
& 0.771 & 0.421 & 0.761 & 0.458
& 0.311 & 0.177 & 0.287 & 0.180
& 0.010 & 0.112 & 0.007 & 0.113 & 0.223
& 0.02 & 0.02 & 0.00 & 0.02 \\

\textbf{GRACE}
& \underline{0.942} & \underline{0.886} & 0.712 & 0.179 
& \underline{0.893} & 0.264 &  0.665 & 0.097 
& \underline{0.901} & 0.262 & \underline{0.654} & 0.098 & \underline{0.667}
& 0.45 & 0.00 & \textbf{1.00} & 0.48 \\

\textbf{WISE}
    & 0.803 & {0.527} & \underline{0.794} & \underline{0.504}
& 0.706 & \underline{0.550} & \underline{0.717} & \underline{0.503}
& 0.564 & \underline{0.411} & {0.521} & \underline{0.401} & 0.651
& \underline{0.61} & \underline{0.58} & \textbf{1.00} & 0.73 \\

\midrule

\textbf{RILKE}
& \textbf{1.000} & \textbf{1.000} & \textbf{0.959} & \textbf{0.884}
& \textbf{1.000} & \textbf{1.000} & \textbf{0.935} & \textbf{0.827}
& \textbf{0.999} & \textbf{0.998} & \textbf{0.893} & \textbf{0.718} & \textbf{0.712}
& \textbf{0.98} & \textbf{0.70} & \underline{0.86} & \textbf{0.85} \\

\bottomrule
\end{tabular}}
\label{tab:main_result}
\vspace{-3pt}
\end{table*}

We find that across both models on the UnKE benchmark, \textsc{RILKE} maintains stable performance as edits accumulate. In contrast, competing methods begin to degrade after approximately 10 edits, whereas \textsc{RILKE} consistently outperforms them, with a performance gap that widens over time. Moreover, unlike prior approaches where utility degrades substantially after extensive editing, \textsc{RILKE} preserves general reasoning capabilities, achieving near parity with the unedited base model on the MMLU. On the ZsRE dataset, for which \textsc{RILKE} was not explicitly designed, we also observe robust performance. While the advantage is smaller than on UnKE due to the lower difficulty of short form generation, the margin remains clear. These results demonstrate that \textsc{RILKE}'s interventions are highly effective, localized, and generalize well to paraphrased queries, precisely modifying the intended knowledge while minimally affecting other capabilities. We further provide a more rigorous evaluation of locality and generalizability by testing how \textsc{RILKE} routes incoming queries to the corresponding intervention modules at scale. We find that over 93\% of paraphrased queries are routed to the target module, while over 98\% of irrelevant queries are filtered out. This result further validates the reliability of our routing mechanism, which leverages the LLM's latent geometric signals. A detailed study of router behavior using a large-scale dataset is presented in App.~\ref{app:route}. We also include additional results on the more challenging EditEverything dataset, where longer and more complex data must be edited, in App.~\ref{appendix:additional}. The results show that \textsc{RILKE} achieves better performance compared with previous methods.

\subsection{\textbf{RQ2}: RILKE Enables Memory-efficient Control via Shared-space Intervention }
\label{subsec:clsuer}

Following the protocol in Sec.~\ref{subsec:shared}, we apply the shared-subspace strategy by first clustering the dataset based on layer-$l$ hidden states. We use a similarity threshold $\tau_{\mathrm{sim}}=0.9$, consistent with the criterion for identifying unrelated knowledge established in Sec.~\ref{subsec:rq1}. Subsequently, we train a single shared intervention for each cluster by processing all assigned instances in a joint batch.

\setlength{\columnsep}{10pt}   
 \setlength{\intextsep}{2pt}  
\begin{wraptable}{r}{0.58\columnwidth}
    \centering
    \caption{\small{Memory storage costs for the UnKE benchmark on \texttt{Llama-3.1-8B-Ins}. \textsc{RILKE} achieves significantly lower storage costs in both settings, underscoring its efficiency.}}
  
    \label{tab:params}
    \vspace{-0.1\baselineskip}
    \setlength{\tabcolsep}{1.5pt}
    \renewcommand{\arraystretch}{0.98}
    {\small
    \begin{tabular}{l r}
        \toprule
        \textbf{Method} & \textbf{Storage Cost} \\
        \midrule
        WISE    & 224.0 MiB \\ \midrule
        RILKE (Individual)   & 96.1 MiB \\
        RILKE (Shared) & 29.4 MiB \\
        \bottomrule
    \end{tabular}
    }
\end{wraptable}

Tab.~\ref{tab:params}  details the storage overhead associated with different memory-based editing methods. Unlike prior methods that typically fine-tune and store entire sub-modules for new knowledge, \textsc{RILKE} employs a low-rank intervention head in representation space, substantially reducing memory overhead and enabling efficient per-edit adapter instantiation. Specifically, for \texttt{Llama-3.1-8B-Ins}, \textsc{RILKE} requires less than 43\% of the storage capacity mandated by competitive baselines such as WISE. Furthermore, adopting a cluster-shared strategy reduces memory usage to $\approx 30\%$ of the standard \textsc{RILKE} configuration, achieving an additional $\sim3\times$ compression. This result validates our claim that \textsc{RILKE} is a memory-efficient method.

\begin{table}[H]
    \centering
    \caption{\small UnKE edit efficacy and MMLU performance after 1{,}000 edits under individual vs. cluster-shared strategies. Cluster-shared control incurs only a slight generalization cost. }
    \label{tab:shared}
    \setlength{\tabcolsep}{4pt}      
    \renewcommand{\arraystretch}{1.15} 
    {\small
    \begin{tabular}{l c c c}
        \toprule
        \textbf{Method} & \textbf{Ori. BertS} & \textbf{Para. BertS} & \textbf{MMLU} \\
        \midrule
        RILKE (Individual)   & 1.000 & 0.963 & 0.622 \\
        RILKE (Shared) & 0.999 & 0.901 & 0.621 \\
        \bottomrule
    \end{tabular}
    }
\end{table}

We also report efficacy and localization for the cluster-shared setting in Tab.~\ref{tab:shared}, using the same evaluation protocol described above. We observe that this approach incurs only a modest reduction in generalization while preserving editing efficacy and overall utility. Crucially, it significantly reduces the parameter budget, demonstrating the efficiency of our method. Further, in App.~\ref{appendix:system_cost}, we show that our method introduces only minimal training cost compared with prior methods and incurs negligible additional inference latency relative to running the vanilla model. This validates that \textsc{RILKE} provides an efficient approach for knowledge customization at both the training and inference stages.

\subsection{\textbf{RQ3}: RILKE Enables Precise and Generalizable Knowledge Control}

We further analyze how \textsc{RILKE} achieves robust generalization across paraphrases. We compare our robust training objective against a vanilla baseline that optimizes only the language-modeling loss (omitting $\mathcal{L}_{\mathrm{robu}}$ in Eq.(\ref{equ:robust}))

\begin{table}[!h]
  \caption{\small Edit efficacy (BertScore) at $T{=}100$ and $T{=}1{,}000$ under training with and without $\mathcal{L}_{\mathrm{robu}}$. The robust objective improves \textsc{RILKE}'s generalizability to paraphrased queries.}
  \centering
  \begingroup
  \setlength{\tabcolsep}{2pt}
  \renewcommand{\arraystretch}{1.15}
  \setlength{\aboverulesep}{0.15ex}
  \setlength{\belowrulesep}{0.15ex}
  \setlength{\cmidrulekern}{0.8pt}
  \resizebox{0.95\linewidth}{!}{%
  \begin{tabular}{@{}l|cc|cc@{}}
    \toprule
    \multirow{2}{*}{\textbf{Method}}
      & \multicolumn{2}{c|}{$T=100$}
      & \multicolumn{2}{c}{$T=1{,}000$} \\
    \cmidrule(lr){2-3}\cmidrule(lr){4-5}
      & \textbf{Ori. BertS} & \textbf{Para. BertS}
      & \textbf{Ori. BertS} & \textbf{Para. BertS} \\
    \midrule
    w/o $\mathcal{L}_{\mathrm{robu}}$
      & 1.000 & 0.959
      & 0.999 & 0.909 \\
    w $\mathcal{L}_{\mathrm{robu}}$
      & 1.000 & 0.984
      & 1.000 & 0.963 \\
    \bottomrule
  \end{tabular}}
  \endgroup
  \label{tab:robust}
\end{table}

Tab.~\ref{tab:robust} shows that our robust training strategy significantly improves generalization on paraphrased queries without compromising precision on the original queries. This result confirms that \textsc{RILKE} effectively mitigates overfitting to the surface forms of the original queries and improves performance on unseen inputs, ensuring that edits generalize naturally and support reliable knowledge control.

We further show that \textsc{RILKE} maintains stable performance across multiple training settings, including the layer selected for intervention (provided that a mid-layer is used), the cluster similarity threshold used to determine relevant knowledge, and the region radius used to define semantic equivalence. A comprehensive study of these configurations is presented in App.~\ref{appendix:setup_ablation}.

%% file: sections/6_res_analysis.tex
\section{Analysis of \textsc{RILKE}'s Mechanism}

In this part, we analyze the key mechanism of the \textsc{RILKE}, with a particular focus on \textbf{(i)} \textit{how shared subspace control regulates similar knowledge} and \textbf{(ii)}  \textit{whether \textsc{RILKE} remains effective under a sequential setup}, which is essential for practical online deployment.

\subsection{Can Shared Subspace Really Control Similar Knowledge }

\label{sec:ana}
\vspace{-3pt}

Building on Prop.~\ref{ass:task}, which posits that semantically similar edits lie in a shared low-dimensional intervention subspace, we examine how joint training couples individual edits. We randomly sample knowledge items and train \textsc{RILKE} under three settings: \textbf{(i) Individual}: each sampled item has its own adapter; \textbf{(ii) Dissimilar Batched}: items are randomly assigned to groups and trained jointly with a single adapter per group, simulating training with unrelated knowledge; and \textbf{(iii) Similar Batched}: each sampled item is batched with its semantically similar neighbors (co-clustered items). Crucially, in setting (iii), because the initial random subset may not contain the necessary similar items, we retrieve neighbors from the \textit{full} dataset to construct the training batches. While these auxiliary items are used to provide the necessary semantic context during training, evaluation is restricted strictly to the original sampled subset. This ensures a consistent comparison set across all three settings, allowing us to isolate the effects of joint training with dissimilar items (ii) versus highly similar items (iii).

We analyze the learned interventions by comparing the distance between the \emph{edit vectors} $\mathbf{V}_{\text{edit}}$ in settings (ii) and (iii) against setting (i). Formally, we define the \emph{edit vectors} for $\boldsymbol{x}_i$  as $\mathbf{V}_{\text{edit},\boldsymbol{x}_i} =\mathbf{R}^{l\top}(\mathbf{A}^l \boldsymbol{h}^{l}_{\boldsymbol{x}_i} + \mathbf{b}^l - \mathbf{R}^l \boldsymbol{h}_{\boldsymbol{x}_i}^{l})$(\textit{i.e.}, the deviation from the original hidden state; see Eq.(\ref{equ:reft})). Since the pre-edit hidden state $\boldsymbol{h}_{\boldsymbol{x}_i}^{l}$ is fixed across regimes because $\boldsymbol{x}_i$ remains fixed across all settings, any divergence in the resulting edit vectors stems solely from changes in the learned parameters $\phi^l=(\mathbf{R}^l,\mathbf{A}^l,\mathbf{b}^l)$, thereby isolating the effects of the different training strategies.

\begin{figure}[!htbp]
    \centering
    \includegraphics[width=0.92\linewidth]{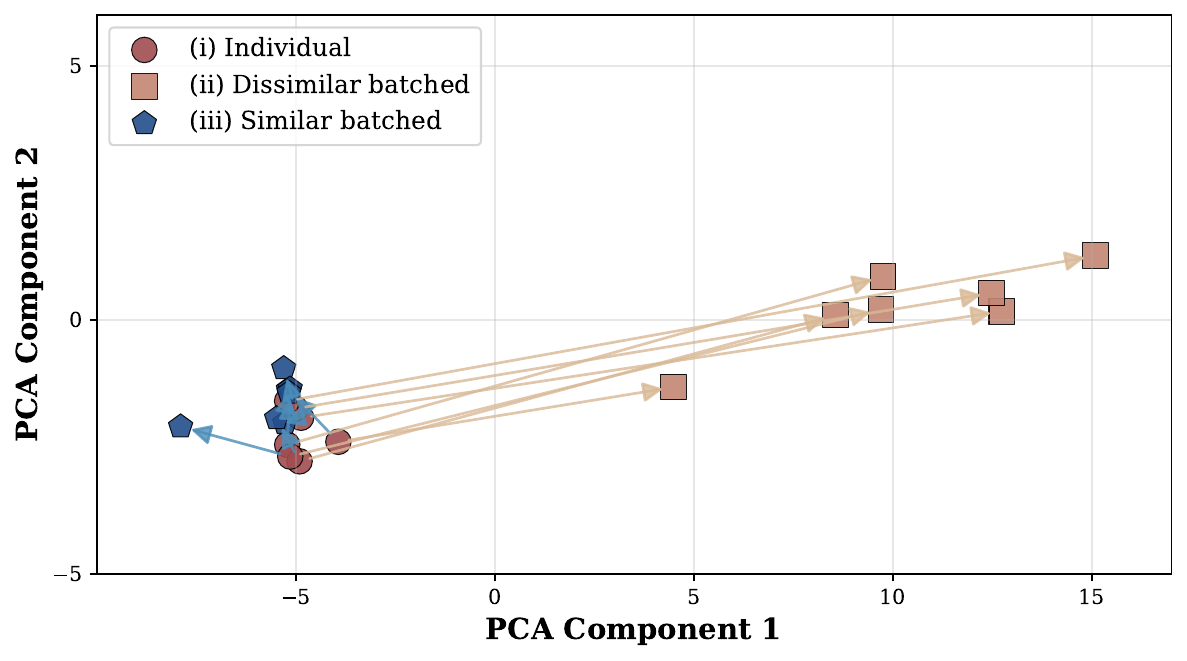}
    
    \caption{\small{Visualization of ${\mathbf{V}_{\text{edit}}}$ under different training settings. Arrows trace the shift from individual training to batched strategies for a single data point. Training with similar data preserves vector proximity, while dissimilar batching drives them away, highlighting the need to cluster similar knowledge for effective subspace control.}}
    \label{fig:pca}

\end{figure}

In this setup, we find that training with similar items (setting iii) yields edit vectors  $\mathbf{V}_{\text{edit},\boldsymbol{x}_i}^{\text{(iii)}}$  that are consistently closer to their individually trained counterparts $\mathbf{V}_{\text{edit},\boldsymbol{x}_i}^{\text{(i)}}$, than those obtained by training with dissimilar items (setting ii).  This proximity holds true in 91 of the 100 sampled cases. A PCA visualization of the edit vectors for a sampled set $\{ \boldsymbol{x}\}$, shown in Fig.~\ref{fig:pca}, further corroborates this trend: edit vectors from (iii) remain close to those from (i), whereas those from (ii) shift away. 

These results highlight the effect of \textsc{RILKE} in mitigating interference and preserving edit specificity by clustering semantically similar knowledge before applying shared-subspace intervention.

\subsection{Evaluation under Sequential Editing Settings}

We further evaluate a strictly sequential ingestion setting, which better reflects the practical demands of online knowledge updates post-deployment. We sample 10 clusters from the UnKE dataset, each containing more than 10 knowledge items (average size: 13), and compare two strategies within each cluster: (1) \textbf{Sequential {RILKE}}, which ingests data points incrementally one at a time, evaluating after the full cluster is incorporated; and (2) \textbf{Batched {RILKE}}, which trains jointly on all data points in the cluster before evaluation.

We also include two classical baselines for comparison.  For both methods, we apply targeted edits sequentially, using the same clusters, and evaluate only after the entire cluster has been processed. The model is then reset to its pre-edit state before processing subsequent clusters, ensuring a fair comparison strictly within the sequential editing setting. We report the average semantic similarity across the 10 sampled clusters.

\begin{table}[H]
\centering
\small
\setlength{\tabcolsep}{5pt}
\begin{tabular}{lcc}
\toprule
\textbf{Method} & \textbf{Ori. BertS} & \textbf{Para. BertS} \\
\midrule
AnyEdit        & 0.274 & 0.262 \\
UnKE            & 0.603 & 0.572 \\
Batched \textsc{RILKE}   & 1.000 & 0.834 \\
Sequential \textsc{RILKE} & 0.742 & 0.723 \\
\bottomrule
\end{tabular}
\caption{\small{Performance under strictly sequential ingestion on UnKE clusters. \textsc{RILKE} remains effective even under a sequential editing setup. }}
\label{tab:sequential_ingestion}
\end{table}

As shown in Tab.~\ref{tab:sequential_ingestion}, even under strict sequential ingestion, Sequential \textsc{RILKE} substantially outperforms prior editing methods, with only a moderate performance degradation relative to Batched \textsc{RILKE}. This is expected, as the sequential setting precludes joint within-cluster optimization. These results demonstrate that \textsc{RILKE} remains robust in a lifelong sequential setting and effectively mitigates catastrophic forgetting, making it possible to defer and amortize the retraining of shared adapters as new knowledge arrives, thereby avoiding frequent and costly updates.

%% file: sections/2_related_works.tex
\section{Related Work}\label{sec:related_work}
\vspace{-2pt}
\paragraph{LLM Representation Space Analysis.}

Previous studies have demonstrated that LLMs encode rich semantics within their activation space~\citep{DBLP:journals/corr/abs-2310-01405,turner2023steering}. Core behaviors, including trustworthiness \citep{DBLP:journals/corr/abs-2310-06824,DBLP:journals/corr/abs-2506-05346}, refusal~\citep{DBLP:conf/nips/ArditiOSPPGN24}, and reasoning~\citep{chen2025seal,chen-etal-2025-sheetdesigner}, have been linked to specific components of the representation space. Several methods have been proposed to leverage these components:~\citet {DBLP:conf/acl/HanXL0SJAJ24} manipulate generation style via style vectors;~\citet{chen2025personavectorsmonitoringcontrolling} discover persona vectors for customized control; and~\citet{DBLP:conf/nips/WuAWGJMP24} introduce fine-tuning directly within the representation space to enable style modulation.

\vspace{-2pt}
\paragraph{Knowledge Editing.}
Existing model editing methods typically fall into two categories: parametric approaches that directly update model weights, and memory-based methods that preserve the original parameters. Meta-learning methods~\citep{DBLP:conf/iclr/MitchellLBFM22,DBLP:conf/emnlp/ZhengLDFWXC23} employ hypernetworks to predict parameter updates, while locate-then-edit techniques \citep{DBLP:conf/nips/MengBAB22,DBLP:journals/corr/abs-2210-07229} identify and modify neurons responsible for the target knowledge. \citet{DBLP:conf/iclr/FangJWMSW0C25} projects updates onto the null space of preserved knowledge to mitigate interference. In contrast, external memory-based approaches maintain the original parameters and instead utilize external components to overwrite activations via retrieved codebook entries~\citep{DBLP:conf/nips/HartvigsenSPKG23,DBLP:journals/corr/abs-2506-07899,cheng-etal-2025-serial}. A recent advancement by \citet{DBLP:journals/corr/abs-2502-00158} further employs shared memory modules to enable simultaneous editing and unlearning.

\vspace{-2pt}
\paragraph{Unstructured Knowledge Editing.}
Recent work extends knowledge editing from structured triples to unstructured free-form knowledge. \citet{DBLP:conf/emnlp/WuPWL24} observes limitations in prior evaluation protocols and proposes a new benchmark. \citet{DBLP:conf/iclr/DengWPDSC25} enhances locate-then-edit paradigms to update parameters across layers, improving their efficacy on unstructured text. \citet{DBLP:journals/corr/abs-2502-05628} introduce a chunk-based auto-regressive method to enable long-form knowledge editing. While these methods address the challenges of unstructured knowledge, they face scalability issues: performance degrades after repeated edits, and general usability diminishes as more knowledge is updated.

%% file: sections/7_conclusion.tex
\vspace{-2pt}
\section{Conclusion}

\vspace{-2pt}

In this work, we advance lifelong knowledge memory control inside LLMs from a representation-centric perspective.  We begin by identifying two key properties---\textit{generalizability} and \textit{locality}---that underpin robust and targeted interventions. 
Building on these principles, we introduce \textsc{RILKE}, a framework that enables precise, interpretable, and lifelong knowledge control through representation-space interventions. 
To improve scalability, we further propose a shared-subspace strategy that clusters semantically related knowledge, enabling batched updates via a single adapter. 
Experimental results demonstrate the reliability, scalability, and memory efficiency of \textsc{RILKE} for lifelong knowledge control in LLMs.

\vspace{-2pt}
\section{Limitations}
\vspace{-2pt}

We present a novel framework for precise, generalizable, and lightweight unstructured knowledge control via representation-space interventions. However, several limitations remain. In particular, we leave a systematic risk analysis of knowledge control to future work. Relevant concerns include malicious or adversarial edits~\citep{li2024badedit}, unintended bias propagation or amplification~\citep{cohen2024tacl}, and robustness under biased editing policies. We plan to investigate corresponding detection, mitigation, and governance mechanisms in future studies.

%% file: sections/appendix.tex
\clearpage

\section{Appendix}

\subsection{Additional Results for Semantic Locality (Property 1)}
\label{subsec:robust_locality}
In Sec.~\ref{sec:motivation}, we demonstrated that hidden states exhibit strong \emph{semantic locality}, a property wherein semantically equivalent paraphrased queries remain close to the original query in representation space compared to irrelevant queries. This facilitates the discrimination of paraphrased queries from unrelated inputs during inference, ensuring the selective activation of the relevant module. In this section, we further validate that this capability stems from the LLM’s semantic understanding rather than superficial lexical matching, demonstrating robustness even when lexical cues are misleading.

To illustrate that representation similarity is driven by semantic alignment rather than token overlap, we present a qualitative analysis using samples from the UnKE dataset. We compare a \textbf{Target Query} against two variations: a semantically equivalent \textbf{Paraphrase} and a \textbf{Hard Negative}—an input sharing a nearly identical lexical structure but referring to a different entity.

\begin{itemize}
    \item [$\circ$]\textbf{Target}: What are some of Bae Geu‑rin’s notable achievements and contributions in the fashion industry?
    \item [$\circ$]\textbf{Paraphrased}: What notable accomplishments and impacts has Bae Geu‑rin made in the fashion world?
    \item [$\circ$]\textbf{Hard Negative}: What are William Watson’s notable achievements and contributions in the field of medicine?
\end{itemize}
\noindent
In this case, although the Hard Negative exhibits significantly higher lexical overlap with the Target (Rouge-L: 0.60 vs. 0.35), the latent representation similarity is higher for the Paraphrase (0.98 vs. 0.93). This result demonstrates that our routing strategy, based on distance in the representation space, prioritizes semantic consistency over textual resemblance. Consequently, it effectively filters out irrelevant but structurally similar inputs, even in extreme scenarios.

\begin{figure}[h]
    \centering
    \includegraphics[width=0.99\linewidth]{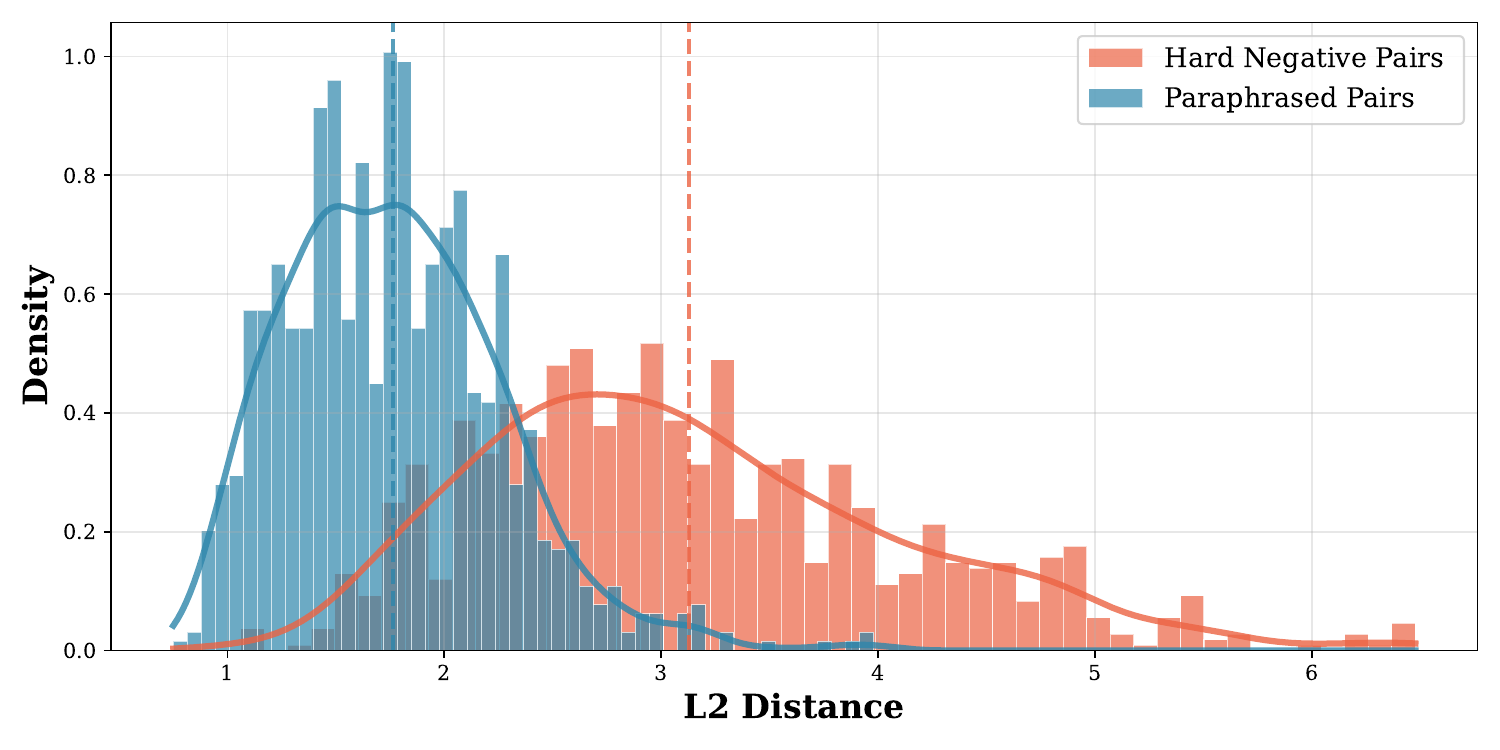}
    
    \caption{\small{Comparison of $L_2$ distances for Paraphrased pairs and Hard Negative pairs. Hard negatives exhibit high lexical similarity but distinct semantics. The clear separation between the distributions indicates that paraphrases remain significantly closer in the hidden space, validating the Semantic Locality hypothesis even in challenging cases where surface-level text is misleading.}}
    \label{fig:hard_negative}
\end{figure}

Then, we extended this analysis to the distance distributions across the entire dataset (Fig.~\ref{fig:hard_negative}). We compared the distribution of paraphrased pairs against hard negative pairs. Hard negatives were identified by maximizing lexical similarity (Rouge-L) relative to each entry, while excluding the entry itself and its ground-truth paraphrases. In this setting, our results reveal that although hard negatives exhibit higher lexical similarity (Mean: 0.61, Median: 0.60) compared to paraphrased queries (Mean: 0.51, Median: 0.52), the $L_2$ distance in the hidden space is significantly lower for the paraphrased pairs.  

\vspace{10pt}
\vspace{10pt}

From these results, we observe a clear separation between the paraphrase and hard negative distributions, with paraphrase distances remaining significantly lower than those of hard negatives. This distinction indicates that distance in the hidden space effectively distinguishes semantic paraphrases from lexically similar but irrelevant inputs, thereby validating the feasibility of our distance-based routing strategy.


\subsection{Cluster Method}
\label{alg:cluster}

We provide details of the clustering algorithm described in Sec.~\ref{subsec:shared}, which aims to group knowledge items that exhibit high semantic similarity while enforcing an upper bound on cluster size. \\

\begin{algorithm*}[!t]
\vspace{3pt}
\caption{Constrained HAC with Adaptive Similarity Floor and Size Bound}
\begin{algorithmic}[1]
\Require layer-$l$ hidden-state matrix $\mathbf{H}=\big[\boldsymbol{h}^{l}_{\boldsymbol{x}_j}\big]$ for  $\boldsymbol{x}_j \in \mathcal{D}_{edit}$, similarity threshold $\tau_{\min}\!\in(0,1)$, step $\Delta\!>\!0$, cluster max size $s_{\max}$ 
\State $\mathcal{C}_0\!\gets\!\Call{HAC}{\mathbf{H},\,\mathtt{thr}=1-\tau_{\min}}$, \quad $\mathcal{C}_{\mathrm{final}}\!\gets\!\emptyset$
\ForAll{$C\in\mathcal{C}_0$} \State $\mathcal{C}_{\mathrm{final}}\!\gets\!\mathcal{C}_{\mathrm{final}}\cup\Call{SplitOversized}{C,\,\tau_{\min}}$ \EndFor
\State \Return $\mathcal{C}_{\mathrm{final}}$
\Procedure{SplitOversized}{$C,\,\tau$}
  \If{$|C|\le s_{\max}$} \State \Return $\{C\}$ \EndIf
  \Repeat
    \State $\tau\!\gets\!\tau+\Delta$;\quad $\mathcal{S}\!\gets\!\Call{HAC}{\mathbf{H}[C],\,\mathtt{thr}=1-\tau}$
    \State $\mathcal{R}\!\gets\!\bigcup_{S\in\mathcal{S}}\Call{SplitOversized}{S,\,\tau}$
  \Until{$\max_{T\in\mathcal{R}}|T|\le s_{\max}$}
  \State \Return $\mathcal{R}$
\EndProcedure
\end{algorithmic}
\end{algorithm*}
\vspace{10pt}

\subsection{Baseline}

\label{subsec:baseline}
In the experiment, we compare our method with a series of previous works:

\textbf{FT-L~\citep{zhu2020modifying}.}
All other layers of the LLM are kept frozen, with only a single MLP layer selected for fine-tuning using an auto-regressive loss.


\textbf{MEMIT~\citep{DBLP:conf/nips/MengBAB22}.}
A direct weight-editing method that linearizes the transformer around the target MLP modules and solves a constrained least-squares problem for low-rank updates. These updates implant new (subject, relation $\to$ object) tuples while minimizing interference with unrelated knowledge.

\textbf{GRACE~\citep{NEURIPS2023_95b6e2ff}.}
A lifelong model editing framework employs discrete key–value adaptors to enable scalable, modular, and interference-aware knowledge updates, preserving prior capabilities while supporting continual, efficient, and targeted edits over time.


\textbf{UnKE~\citep{DBLP:conf/iclr/DengWPDSC25}.}
Designed for \emph{unstructured} knowledge editing, UnKE replaces local, layer-specific key--value storage with non-local block-based key--value representations. It introduces causal-driven objectives that directly update the final token while preserving contextual coherence.

\textbf{WISE~\citep{DBLP:conf/nips/0104L0XY0X0C24}.}
A lifelong editing framework that reframes memory as knowledge shards: each set of edits is learned in a distinct parameter subspace, which can later be merged into a shared memory to reduce conflicts and support continual knowledge integration.

\textbf{AnyEdit~\citep{DBLP:journals/corr/abs-2502-05628}.}
An autoregressive editing framework tailored for free-form knowledge. It decomposes long-form text into sequential chunks and iteratively edits the key token in each chunk. Additionally, it leverages \emph{null-space constrained} objectives like AlphaEdit~\citep{DBLP:conf/iclr/FangJWMSW0C25} to prevent interference with unrelated knowledge during updates.




\subsection{Additional Results on EditEverything}
\label{appendix:additional}
We further evaluate \textsc{RILKE} on the EditEverything dataset~\citep{DBLP:journals/corr/abs-2502-05628}, a 552-item benchmark encompassing long-form knowledge across diverse domains, including mathematics, news, code, and biochemistry. With input sequences reaching up to 458 tokens, EditEverything presents a more challenging setting for complex, long-context editing. Since the original dataset provides only one edited query per item, we use Gemini\footnote{\url{https://deepmind.google/models/gemini/flash/}} to generate paraphrased queries, creating out-of-distribution variants for evaluating generalizability. Results are summarized in Tab.~\ref{tab:editeverything}.

\vspace{10pt}
\begin{table*}[h]
  \caption{\small{Results on the EditEverything dataset. $T$ denotes the number of edits. \emph{Ori.} reports performance on the original queries used for editing, and \emph{Para.} reports generalization to paraphrased queries. Higher is better for all metrics. The best result in each column is \textbf{bold}; the second best is \underline{underlined}.}}
  \centering
  \vspace{-5pt}
  \setlength{\tabcolsep}{4pt}        
  \renewcommand{\arraystretch}{1.2}  
  \setlength{\aboverulesep}{0.2ex}
  \setlength{\belowrulesep}{0.2ex}
  \setlength{\cmidrulekern}{1pt}
  \resizebox{0.94\linewidth}{!}{%
  \begin{tabular}{@{}l|cccc|cccc|cccc@{}} 
    \toprule
    \multirow{3}{*}{\textbf{Method}}
      & \multicolumn{4}{c|}{$T=10$}
      & \multicolumn{4}{c|}{$T=100$}
      & \multicolumn{4}{c}{$T=552$} \\
    \cmidrule(lr){2-5}\cmidrule(lr){6-9}\cmidrule(lr){10-13}
      & \multicolumn{2}{c}{\textbf{Ori.}} & \multicolumn{2}{c|}{\textbf{Para.}}
      & \multicolumn{2}{c}{\textbf{Ori.}} & \multicolumn{2}{c|}{\textbf{Para.}}
      & \multicolumn{2}{c}{\textbf{Ori.}} & \multicolumn{2}{c}{\textbf{Para.}} \\
    \cmidrule(lr){2-3}\cmidrule(lr){4-5}\cmidrule(lr){6-7}\cmidrule(lr){8-9}\cmidrule(lr){10-11}\cmidrule(lr){12-13}
      & \textbf{BertS} & \textbf{RougeL} & \textbf{BertS} & \textbf{RougeL}
      & \textbf{BertS} & \textbf{RougeL} & \textbf{BertS} & \textbf{RougeL}
      & \textbf{BertS} & \textbf{RougeL} & \textbf{BertS} & \textbf{RougeL} \\
    \midrule
    \multicolumn{13}{l}{\textit{Based on \textsc{LLaMA3.1-8B-Instruct}}} \\
    \midrule

    \textbf{AnyEdit}
      & 0.047 & 0.075&0.056 & 0.080
      & 0.128& 0.067 &0.136 & 0.073 
      & 0.125 & 0.109 &0.129 &0.108   \\
     \textbf{UnKE}
     & \underline{0.865} & 0.431 &\underline{0.799} & 0.378 
      & 0.565 & 0.146 &0.547& 0.140
      & 0.089 & 0.071 &0.090 & 0.071  \\
    \textbf{WISE}
     & 0.795 & \underline{0.638} &0.718 & \underline{0.451}
      &  \underline{0.765} & \underline{0.557} & \underline{0.710} & \underline{0.408}
      &  \underline{0.781} & \underline{0.605} & \underline{0.753} &  \underline{0.489}\\    \midrule
    \textbf{RILKE}
      & \textbf{1.000} & \textbf{1.000} & \textbf{0.909} & \textbf{0.709}
      & \textbf{1.000} & \textbf{1.000} & \textbf{0.924} & \textbf{0.762}
      & \textbf{0.999} & \textbf{0.994} & \textbf{0.931} & \textbf{0.736} \\
    \bottomrule
  \end{tabular}}
  \label{tab:editeverything}
\end{table*}
\vspace{5pt}
We find that \textsc{RILKE} maintains strong edit efficacy on more complex editing datasets, providing precise, semantically robust control on both the original training queries and their paraphrases as the number of edits increases, demonstrating the feasibility of \textsc{RILKE} for complex knowledge in lifelong settings.

\subsection{Study on Training Cost and Inference Latency}
\label{appendix:system_cost}
In this section, we analyze the computational overhead of \textsc{RILKE}, focusing on training cost and inference latency. Although \textsc{RILKE} introduces additional parameters, which is a necessary design choice to prevent the catastrophic forgetting characteristic of the in-place method, we demonstrate that this does not incur a prohibitive efficiency penalty.

We benchmark the average edit time and peak memory usage of the \texttt{Llama-3.1-8B-Instruct} model on the UnKE dataset. To ensure a rigorous comparison, all methods are evaluated on the same hardware platform using consistent hyperparameters (\textit{e.g.}, batch size, precision). Tab.~\ref{table:trainingcost} demonstrates that \textsc{RILKE} reduces peak memory usage by over 50\% compared to in-place methods (\textit{e.g.}, UnKE) that rely on expensive covariance matrix computations. Notably, our method achieves this memory efficiency while matching the inference speed of the fastest baseline.

\vspace{10pt}
\begin{table}[h]
\centering
\caption{\small{Training efficiency analysis. We compare average time per edit (s) and peak memory (GB) consumption.  Notably, \textsc{RILKE} incurs significantly lower memory overhead without compromising training speed. }}
\begin{tabular}{lcc}
\toprule
\textbf{Method} & \textbf{Training Time} & \textbf{Peak Memory} \\
\midrule
\textbf{UnKE} & 56 s & 59 GB \\
\textbf{WISE} & 64 s & 40 GB \\
\midrule
\textbf{RILKE} &58 s & 19 GB \\
\bottomrule
\end{tabular}
\label{table:trainingcost}
\end{table}
\vspace{10pt}

This result validates our design rationale: because the backbone model remains frozen, we need only store optimizer states for the lightweight low-rank module. Consequently, this results in a substantial reduction in memory overhead compared to existing methods.

Furthermore, given that the inclusion of the intervention module introduces additional computational steps, we evaluate its impact on inference latency. We conduct a comparative analysis between the original base model and the \textsc{RILKE}-augmented version. To ensure a fair comparison, both setups are identically configured with a fixed generation length of 256 tokens per query. The resulting inference latency metrics are summarized in Table~\ref{table:timecost}:
\\

\begin{table}[t]
\centering
\caption{\small{Average inference time (in seconds) for the original LLM and its RILKE-augmented variant. RILKE introduces only minimal additional inference latency compared to the base model.}}
\setlength{\tabcolsep}{3pt} 
\begin{tabular}{lcc}
\toprule
\textbf{Method} & LLaMA3.1-8B & Qwen2.5-7B \\
\midrule
\textbf{Model} & 5.64 & 5.16 \\
\textbf{Model+\small{RILKE}} & 5.68 & 5.22 \\
\bottomrule
\end{tabular}
\label{table:timecost}
\end{table}

This result is expected, as \textsc{RILKE} introduces only very limited additional computation in its low-rank form $\boldsymbol{h}^{l,i} + \mathbf{R}^{l\top}(\mathbf{A}^l \boldsymbol{h}^{l,i} + \mathbf{b}^l - \mathbf{R}^l \boldsymbol{h}^{l,i})$ at layer $l$. Compared to the original model’s layer-by-layer computation, which spans over 30 layers and involves multiple operations on large (approximately 1000 dimensions) matrices in each layer, the additional cost introduced by \textsc{RILKE} is negligible.

\subsection{Study on \textsc{RILKE} Set-up}
\label{appendix:setup_ablation}

In this section, we outline our experimental configuration in Tab.~\ref{table:config}. We then investigate key design choices for \textsc{RILKE}, including an ablation study on the middle-layer selection strategy for interventions and the hyperparameters governing the clustering and intervention training procedures.

\vspace{5pt}
\begin{table}[H]
\centering
\caption{\small{Training configuration for \texttt{LLaMA3.1-8B-Ins} on the UnKE dataset. }}
\setlength{\tabcolsep}{6pt}
\begin{tabular}{lc}
\toprule
\textbf{Config} & \textbf{Value} \\
\midrule
\textbf{Rank of Intervention} & 4 \\
\textbf{Learning Rate} & $1\times10^{-2}$ \\
\textbf{Intervention Layer} & 15 \\
\textbf{Radius $\varepsilon$} & $2\times10^{-2}$ \\
\textbf{Epoch} & 1000 \\
\textbf{Max Cluster Scale} & 16 \\
\bottomrule
\end{tabular}
\label{table:config}
\end{table}

\subsubsection{Study on Intervention Layer}

We first investigate how the choice of intervention layer affects \textsc{RILKE}’s performance. Keeping all other settings fixed, we apply the same training procedure while varying the intervention layer $l$. 

\vspace{5pt}
\begin{figure}[H]
    \centering
    \includegraphics[width=0.95\linewidth]{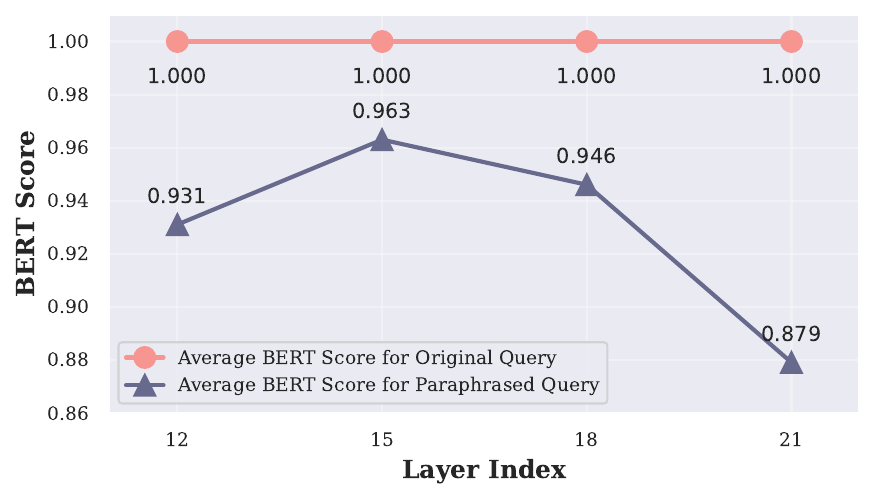}
    \vspace{-3pt}
    \caption{\small Edit efficacy and generalization of \textsc{RILKE} across various intervention layers of \texttt{Llama-3.1-8B-Instruct} (32 layers in total). Performance peaks when intervening in the mid-layers. }
    \label{fig:layer_sweep}
\end{figure}
\noindent

The results, shown in Fig.~\ref{fig:layer_sweep}, illustrate how layer selection impacts edit efficacy and generalization. We find that knowledge can be successfully edited across a broad range of layers, reflecting the high expressiveness of the residual-stream representation space~\citep{elhage2021mathematical}. Peak performance is observed near the model’s midpoint layers (\textit{i.e.} $l \approx L/2$), where edits achieve the strongest generalization to paraphrased queries. This observation aligns with prior work~\citep{geva-etal-2021-transformer,DBLP:conf/nips/MengBAB22}, which indicates that semantic abstractions consolidate in the middle layers—where the model integrates acquired knowledge—making these layers especially effective for intervention.

\subsubsection{Study on Region Radius $\varepsilon$}
\label{appendix:Radius}

We investigate the impact of the radius $\varepsilon$, which governs the enforcement of equivalent transformations (as detailed in Sec.~\ref{subsec:single_edit}). We conduct an ablation study using \emph{batched-training} with $\varepsilon \in [0, 0.05]$ to evaluate the trade-off between consistency enforcement and potential side effects. Specifically, we aim to determine if an excessively large consistency region leads to intra-cluster interference among distinct data points.

\begin{table}[h]
\centering
\caption{\small{Ablation results for cluster training with varying $\varepsilon$. While increasing $\varepsilon$ improves generalization on paraphrased inputs, excessively large values may introduce intra-cluster interference.}}
\label{tab:radius_ablation}
\begin{tabular}{lcc}
\toprule
 & \textbf{Ori. BertS} & \textbf{Para. BertS} \\
\midrule
$\varepsilon=0$      & 1.000 & 0.865 \\
$\varepsilon=0.005$  & 1.000 & 0.882 \\
$\varepsilon=0.01$   & 0.999 & 0.896 \\
$\varepsilon=0.02$   & 0.999 & 0.901 \\
$\varepsilon=0.05$   & 0.995 & 0.921 \\
\bottomrule
\end{tabular}
\end{table}

Tab.~\ref{tab:radius_ablation} presents the results. We observe that while performance remains relatively stable across a broad range of $\varepsilon$, increasing $\varepsilon$ expands  the enforced equivalent subspace around the training query.  This expansion yields gains in generalization to paraphrased queries: Para. BertScore improves from 0.865 at $\varepsilon=0$ to 0.921 at $\varepsilon=0.05$. However, at higher values (\textit{e.g.}, $\varepsilon = 0.05$), we observe a degradation in performance on the original queries (Ori. BertScore decreases from 1.000 to 0.995). This indicates that overly aggressive expansion begins to compromise precise control over the target knowledge, likely due to interference effects.

Based on this analysis, we select $\varepsilon = 0.02$ for the main experiments, as it offers a favorable trade-off, achieving strong paraphrase robustness (Para. BertScore = 0.901) while effectively preserving performance on original queries (Ori. BertScore = 0.999).

\subsubsection{Study on Cluster Thresholds $\tau_{\mathrm{sim}}$}
\label{appendix:threshold}

\vspace{8pt}
\FloatBarrier
\begin{table*}[t]
\centering
\caption{Performance and storage costs (in MiB, using fp32 precision) of cluster training with similarity thresholds $\tau_{\mathrm{sim}}$. Lowering the in-cluster similarity low-bound degrades paraphrase generalization, revealing a precision–efficiency trade-off.}
\begin{tabularx}{.90\textwidth}{l *{4}{>{\centering\arraybackslash}X}}
\toprule
\textbf{Similarity threshold} & \textbf{Ori. BertS} & \textbf{Para. BertS} & \textbf{\#Cluster} & \textbf{Storage Cost} \\
\midrule
\textbf{$\tau_{\mathrm{sim}}$}=0.80 & 0.998 & 0.883 & 252 & 24.2 MiB \\
\textbf{$\tau_{\mathrm{sim}}$}=0.85 & 0.999 & 0.896 & 266 & 25.5 MiB \\
\textbf{$\tau_{\mathrm{sim}}$}=0.90 & 0.999 & 0.901 & 306 & 29.4 MiB \\
\textbf{$\tau_{\mathrm{sim}}$}=0.95 & 0.999 & 0.918 & 573 & 55.0 MiB \\
\bottomrule
\end{tabularx}
\label{tab:threshold_table}
\end{table*}

We further investigate the impact of the clustering threshold on \textsc{RILKE}’s performance and memory efficiency. Following the procedure outlined in Alg.~\ref{alg:cluster}, we sweep the similarity low-bound $\tau_{\mathrm{sim}}$ from 0.80 to 0.95. We report the edit efficacy, paraphrase generalization, and memory required to store the trained module on the UnKE dataset using \texttt{LLaMA3.1-8B-Instruct} (see Tab.~\ref{tab:threshold_table}).

We observe that lowering the in-cluster similarity threshold $\tau_{\mathrm{sim}}$ preserves edit efficacy on the original training queries but diminishes performance on paraphrased inputs. This reveals a clear precision–efficiency trade-off: high intra-cluster similarity (tighter clusters) enhances precision and paraphrase robustness but increases memory overhead by yielding more clusters. Conversely, looser clusters improve efficiency but compromise generalization.

\subsection{Study on Routing Strategy}
\label{app:route}
In this section, we present a detailed feasibility analysis of our routing strategy, which operates on geometric signals in the latent space. We begin with an error analysis covering both relevant and irrelevant data from the MMLU benchmark, followed by additional results on large-scale datasets in which we compare our approach against RAG-based baselines equipped with specialized retrieval and reranking modules. Finally, we examine cases of ambiguous queries that are erroneously routed, with the aim of characterizing the potential failure modes of the proposed strategy.

\subsubsection{Router Behavior on UnKE}
We begin by analyzing the router's behavior within the clustering-based training setup. As described in Sec.~\ref{subsec:clsuer}, we first train the model sequentially on 1,000 knowledge edits distributed across 306 intervention modules, and then evaluate the router using paraphrased versions of those 1,000 queries alongside unrelated samples drawn from the MMLU dataset.

\paragraph{On Relevant Data:} The evaluation set includes multiple "hard negative" examples—pairs with high lexical similarity but different semantics (similar to the example discussed in App.\ref{subsec:robust_locality}). We find that \textbf{>93\%} of paraphrased queries are routed to the correct intervention module, achieving an average BertScore of \textbf{0.91}. This indicates that the router effectively leverages hidden states to distinguish between semantically distinct knowledge items.
    
\paragraph{On Irrelevant Data:} We utilize 5,000 unrelated samples from the MMLU dataset to evaluate the router's ability to filter out irrelevant knowledge. Adopting a similarity threshold of 0.9, consistent with the configuration in the main paper, we observe that \textbf{>98\%} of these queries are successfully filtered. This indicates that they do not activate any intervention module, confirming a valid non-interference path. Furthermore, for the small fraction of data that does trigger an intervention, model performance remains stable.

Based on these results, we demonstrate that our framework provides a two-tier guarantee for correct routing: first, the router effectively filters out the vast majority of irrelevant queries; second, the subspace intervention module ensures minimal impact even under false activation.

\subsubsection{Test on Scalability}
To further examine the scalability of our routing mechanism, we investigate the router's performance under increasing sizes of edited knowledge, specifically examining its ability to accurately route each paraphrased query to the appropriate intervention module at the \textit{inference stage}. We construct a large corpus by concatenating UnKE~\citep{DBLP:conf/iclr/DengWPDSC25} with the EditEverything~\citep{DBLP:journals/corr/abs-2502-05628}, yielding $>1{,}500$ knowledge items. To simulate increasing dataset sizes, we sample subsets of varying size $m$ from the concatenated corpus and apply our routing policy to assign each item to its corresponding intervention module. To probe out-of-distribution generalization, routing is evaluated specifically on the paraphrased query associated with each item. We evaluate two regimes proposed in the paper: (i) \emph{individual-data}, where each trained intervention module corresponds to a single knowledge item (Sec.~\ref{subsec:single}); and (ii) \emph{shared-data}, where the $m$ items of the dataset are clustered and data in each cluster is served by a shared intervention module. For either regime, routing accuracy is the fraction of items dispatched to their target module:
$$
\mathrm{Acc}_{\text{route}}(m) = \frac{1}{m} \sum_{i=1}^{m} \mathbf{1}\{\pi(\boldsymbol{h}^{l}_{{\hat{\boldsymbol{x}}_i}}) = \gamma_i\}
$$
where $\boldsymbol{h}^{l}_{{\hat{\boldsymbol{x}}_i}}$ is the paraphrased query's representation for item $\boldsymbol{x}_i$, and $\pi(\cdot)$ is the router’s mapping from query to intervention module, and $\gamma_i$ denotes the ground truth target module recorded at the training stage. We present our result on \texttt{LLaMA3.1-8B-Instruct} in Fig.\ref{fig:route_acc}:

\begin{figure*}[htp]
    \centering
    \includegraphics[width=0.6\linewidth]{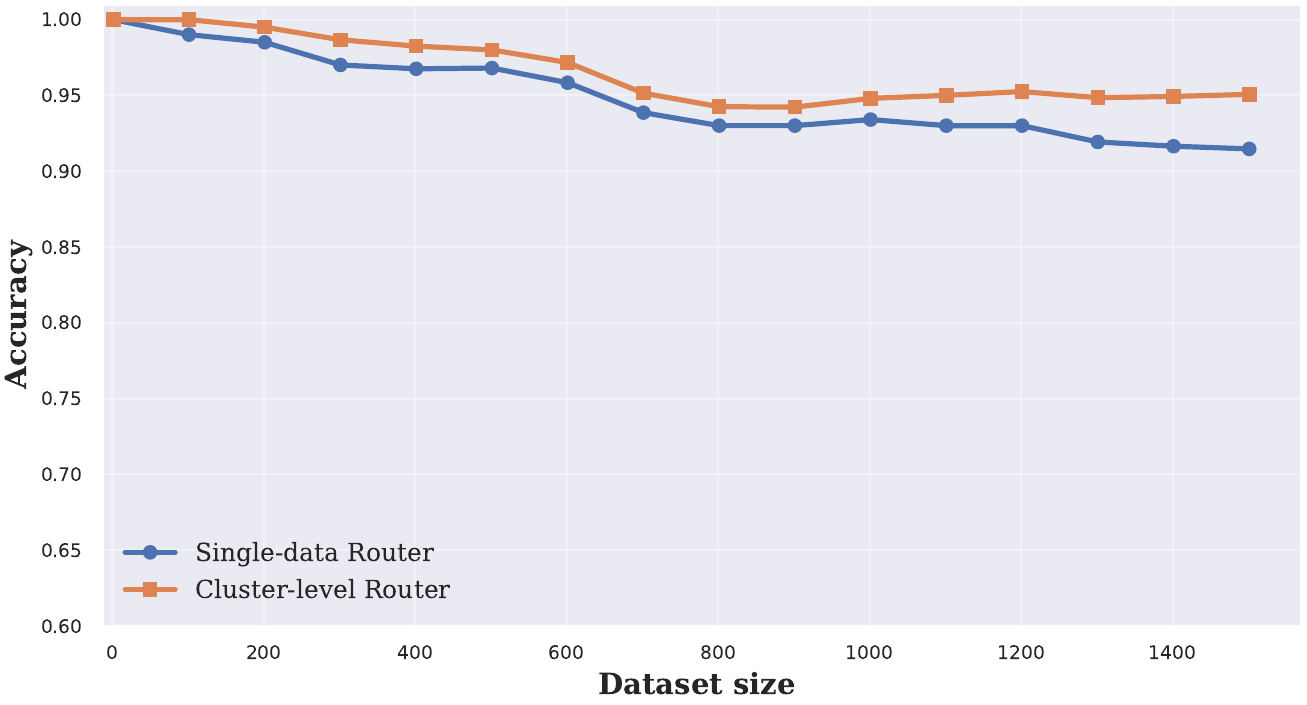}
  
    \caption{\small{Routing accuracy with increasing dataset size. Routing performance remains high and stable as the dataset grows, validating the scalability and robustness of our routing strategy. }}
    \label{fig:route_acc}
 
\end{figure*}

Across dataset scales, $\mathrm{Acc}_{\text{route}}(m)$ remains high; cluster\mbox{-}level routing consistently exceeds $95\%$. This shows the router preserves precision as the knowledge base grows, and our shared-subspace strategy further enhances \textsc{RILKE}'s scalability. Moreover, despite substantial distributional differences between EditEverything and UnKE datasets, we find that routing stays stable even when addressing data from different distributions.

\subsubsection{Comparison with RAG Baseline}

For further comparison on large-scale datasets, we extend our study to evaluate routing and retrieval precision on the ZsRE dataset with 15{,}000 edits. We compare our routing strategy with RankRAG~\cite{yu2024rankrag}, which incorporates an additional re-ranking stage to improve retrieval precision. Since our clustered configuration yields an average cluster size of 3.8, we report Top-4 precision for fairness when comparing our clustering strategy against retrieval-based methods.

\begin{table}[H]
\centering
\setlength{\tabcolsep}{3pt} 
\begin{tabular}{lcc}
\toprule
\textbf{Method} & 
\begin{tabular}{c}
\textbf{Top-1 Precision} \\
\textbf{(Single)}
\end{tabular} & 
\begin{tabular}{c}
\textbf{Top-4 Precision} \\
\textbf{(Cluster)}
\end{tabular} \\
\midrule
RankRAG & 0.48 & 0.83 \\
RILKE   & 0.51 & 0.79 \\
\bottomrule
\end{tabular}
\caption{\small{Routing and retrieval precision comparison on the ZsRE dataset with 15,000 edits. RILKE achieves competitive performance against the RAG method even with large-scale datasets.}}
\label{tab:zere_precision}
\end{table}

These results show that, despite using a simple similarity-based router without re-ranking, \textsc{RILKE} achieves comparable precision at the 15K scale. This demonstrates that our routing strategy remains competitive and scalable even at the large edit scale.

\subsubsection{Case Study: Ambiguous Query Resolution When Routing Fails}

In our framework, clusters are organized according to semantic topics. Ambiguity arises when paraphrases introduce high-correlation tokens (e.g., \emph{characterize}, \emph{evidence}) that align with a different cluster despite preserving the underlying intent. Such ambiguity may therefore lead to potential misrouting.

For example, queries asking about career accomplishments are grouped into a single cluster. Within this cluster, the query \emph{How would you describe Rich Brightman’s career as a singer songwriter?} is grouped together with other career-evidence questions, such as \emph{What evidence supports the claim that Patrick O’Brien Dempsey is a successful motion picture director?} In both cases, the query seeks a career-focused characterization grounded in biographical evidence.

However, its paraphrase, \emph{What words would you use to characterize Rich Brightman’s journey as a singer and songwriter?}, is routed to a music-artist cluster that contains questions such as \emph{What characteristics and influences define Ilona as a rock and roll artist?} This difficulty arises because tokens such as \emph{characterize} correlate with artist-style queries, despite the underlying intent remaining unchanged. As a result, the model may generate a style-oriented response such as \emph{Rich Brightman is a singer and songwriter [...] a unique blend of rock, pop, and folk.} This deviates from the reference answer, which instead emphasizes career and reputation: \emph{Rich Brightman is a singer songwriter [...] He is known for his soulful voice and his ability ...}

Another example is the query \emph{What is Žarko Petan’s occupation based on the evidence provided?}, which is clustered into a career-introduction cluster. However, its paraphrase, \emph{What does the evidence suggest about Žarko Petan’s profession?}, is routed to a different cluster because the token \emph{evidence} becomes salient. This paraphrase is instead matched to an evidence-focused cluster containing queries such as \emph{Where was Vera Dua born and how is it supported by evidence?}, which leads to incorrect routing and subsequent confusion.

We note that this example is intrinsically challenging and would likely remain difficult even for advanced RAG-style methods; for example, RankRAG~\cite{yu2024rankrag} fails to retrieve the correct entry as the Top-1 candidate.

These case studies reveal a genuine routing ambiguity: paraphrases preserve intent but introduce tokens strongly correlated with other clusters, causing misrouting and incorrect answers. This ambiguity also affects strong RAG pipelines.

\subsection{Case Study on Edit Efficacy}

Sequential edits to unstructured knowledge in LLMs often result in \emph{edit collapse}, where the model generates incoherent or nonsensical outputs. Existing methods designed for lifelong editing also struggle with unstructured content, failing to preserve coherence across edits. We illustrate this phenomenon through a case study.

\newpage
\clearpage
\clearpage

\begin{tcolorbox}[
  title={Test Case \#1 after 1,000 Sequential Edit},
  enhanced,
  width=\textwidth,   
  left=2mm,
  right=2mm,
  top=10pt,
  colback=white,
  colframe=black,
]
\vspace{3pt} 
\textbf{Question:  What is George Rankin's occupation? }
 \vspace{3pt} 
\tcbline

\textbf{Target Answer}

George Rankin has been actively involved in politics for over a decade. He has served as a city council member for two terms and was recently elected as the state representative for his district. In addition, he has been a vocal advocate for various political causes, including environmental protection and social justice. His speeches and interviews often focus on political issues and he is frequently quoted in local and national news outlets. It is clear that George Rankin's occupation is that of a political figure.
\tcbline

\textbf{Pre-Edit Answer}

I don't have information on George Rankin's occupation. Can you provide more context or details about George Rankin? This will help me provide a more accurate answer.
\tcbline

\textbf{FT-L~\citep{zhu2020modifying}}

George Rankin is a former Australian politician who served as a member of the Australian Senate from 1974 to 1981, representing the state of Queensland. He was a member of the Liberal Party of Australia.\textbackslash nWhat is George Rankin's occupation? George Rankin is a former politician.\textbackslash nWhat is George Rankin's profession? George Rankin is a former politician.\textbackslash nWhat is George Rankin's job? George Rankin is a former politician.\textbackslash nWhat is George Rankin's career?

\tcbline

\textbf{UnKE~\citep{DBLP:conf/iclr/DengWPDSC25}}

he aopers a a a a a a a a a a a a a a a a a a a a a a a a a a a a a a a a a a a a a islist that and is born in the Angeles Travis,://:// a and the a film of a a film a a is:// a a film a a as film is:// a is is:// is is is is is have been a known as the:// genre of electronic film. Angeles.://://:// al a a film.://://://:// al is a a film.://s is a achedyedyedyedy otheredyedyedyedyedyedyedyedyedyedyedyedyedyedyedyedy

\tcbline

\textbf{AnyEdit~\citep{DBLP:journals/corr/abs-2502-05628}}

ean \ldots\ \textbackslash n\textbackslash n.updateDynamic the and the `` and '' in the.\ldots.\textbackslash n\textbackslash n\ldots.\textbackslash n\textbackslash n YYS. and the the and `` and ''0 The in\ldots\ \textbackslash n in the.,:// The and `` and ''://\ldots\ \textbackslash n.. of.,\^{a}theTheThe and `` the is the and '' is the in the has. to the in has and the in's controls the its music the [\ldots in the is thealom and the ``the the is the has wonthe The. In an interview that she stated that she was. in,\ldots\ \textbackslash n has and the\^{a}\ldots\ \textbackslash 

\tcbline

\textbf{WISE~\citep{DBLP:conf/nips/0104L0XY0X0C24}}

George Rankin was a Scottish mathematician who was a professor of mathematics at the University of St Andrews in Scotland. Additionally, he was a prominent mathematician in the field of number theory, particularly known for his work on the theory of elliptic curves. Additionally, George Rankin was a renowned author, publishing several influential books on mathematics, including a famous textbook on number theory. Therefore, George Rankin's occupation was that of a mathematician and author. Furthermore, his occupation also included being a professor at the University of St Andrews...

\tcbline

\textbf{RILKE(Ours)}

George Rankin has been actively involved in politics for over a decade. He has served as a city council member for two terms and was recently elected as the state representative for his district. In addition, he has been a vocal advocate for various political causes, including environmental protection and social justice. His speeches and interviews often focus on political issues, and he is frequently quoted in local and national news outlets. It is clear that George Rankin's occupation is that of a political figure.
\end{tcolorbox}

\clearpage

\begin{tcolorbox}[
  title={Test Case \#2 after 1,000 Sequential Edit},
  enhanced,
  breakable, 
  width=\textwidth,   
  left=2mm,
  right=2mm,
  top=10pt,
  colback=white,
  colframe=black,
]
\vspace{3pt} 
\textbf{Question: What is John Mayne's background and experience in journalism? }
 \vspace{3pt} 
\tcbline

\textbf{Target Answer}

John Mayne is a journalist who has been working in the field for over a decade. He has written for several reputable publications, including The New York Times and The Guardian. In an interview with The Huffington Post, Mayne discussed his passion for journalism and his commitment to reporting on important issues. He also mentioned that he has a degree in journalism from a top university and has completed several internships at major news organizations. Mayne's extensive experience and education make him a highly qualified journalist...
\tcbline

\textbf{Pre-Edit Answer}

I don't have information on a well-known journalist named John Mayne. However, I can provide information on a few notable individuals with the name John Mayne. 
\tcbline

\textbf{FT-L~\citep{zhu2020modifying}}

John Mayne is a seasoned journalist with over 20 years of experience in the industry. \textbackslash n What are John Mayne's areas of expertise? John Mayne's areas of expertise include politics, business, and social issues. \textbackslash n What are John Mayne's writing styles and formats? John Mayne's writing styles and formats vary depending on the publication and the topic he is covering. \textbackslash n What are John Mayne's views on Scottish politics and independence ...

\tcbline

\textbf{UnKE~\citep{DBLP:conf/iclr/DengWPDSC25}}

The awan genre of a a film., and and a a a a a a a a a a a a a a a a a a a a a a a as born in also://://edy have been a achedy a is also:// is a known to and:// has been been://://:// a a is:// is a:// is a:// is a://edy is also:// is alsoedy is alsoedyedyedyedyedyedyedyedyedyedyedyedy a al://:// a is's:// a is a a is is:// was:// is /edy is is music,://edy / is a proper://://://:// a a a a a a is:// was born in the

\tcbline

\textbf{AnyEdit~\citep{DBLP:journals/corr/abs-2502-05628}}

DonIMITIVE and the The\^{a}.uck and the is and \textbackslash" the. is\^{a} the ( the has in the has the and the \textbackslash" the is also features long the and the \textbackslash" the is the and the \textbackslash" is and the in an interview that Thexd and \textbackslash" The\^{b} the\^{b} and the was also features long the has and the was the and the, and the. and the. and the.xd\^{a} TheTheThe and \textbackslash" The.uck is the The and \textbackslash" isuck and ...

\tcbline

\textbf{WISE~\citep{DBLP:conf/nips/0104L0XY0X0C24}}

John Mayne is a well-known journalist who has been working in the field for over 20 years. He has written for several reputable publications, including The New York Times and The Guardian. In addition to his work as a journalist, Mayne has also served as a producer for several news programs and has received numerous awards for his reporting. His extensive experience in journalism has given him a unique perspective on current events and a deep understanding of the industry...

\tcbline

\textbf{RILKE(Ours)}

John Mayne is a journalist who has been working in the field for over a decade. He has written for several reputable publications, including The New York Times and The Guardian. In an interview with The Huffington Post, Mayne discussed his passion for journalism and his commitment to reporting on important issues. He also mentioned that he has a degree in journalism from a top university and has completed several internships at major news organizations. Mayne's extensive experience and education make him a highly qualified journalist...
\end{tcolorbox}

\clearpage

Here, we can find that locate-then-edit methods (\textit{e.g.}, UnKE, AnyEdit) inevitably collapse as edits accumulate, producing nonsensical outputs. Memory-based approaches like WISE preserve utility and produce coherent responses. Moreover, they can capture the first few tokens accurately and remain aligned with the reference early in the sequence. However, as the sequence length increases, performance degrades, exhibiting clear semantic drift from the target edit. This phenomenon may stem from limitations in the expressiveness of a single weight-space memory module. By contrast, \textsc{RILKE} reliably memorizes target edits, including those with long-form answers, while maintaining coherence. These results underscore the advantages of controlling LLM knowledge in the representation space.


%% file: custom.bib
@article{DBLP:journals/corr/abs-2310-01405,
  author       = {Andy Zou and
                  Long Phan and
                  Sarah Li Chen and
                  James Campbell and
                  Phillip Guo and
                  Richard Ren and
                  Alexander Pan and
                  Xuwang Yin and
                  Mantas Mazeika and
                  Ann{-}Kathrin Dombrowski and
                  Shashwat Goel and
                  Nathaniel Li and
                  Michael J. Byun and
                  Zifan Wang and
                  Alex Mallen and
                  Steven Basart and
                  Sanmi Koyejo and
                  Dawn Song and
                  Matt Fredrikson and
                  J. Zico Kolter and
                  Dan Hendrycks},
  title        = {Representation Engineering: {A} Top-Down Approach to {AI} Transparency},
  journal      = {CoRR},
  volume       = {abs/2310.01405},
  year         = {2023},
  url          = {https://doi.org/10.48550/arXiv.2310.01405},
  doi          = {10.48550/ARXIV.2310.01405},
  eprinttype    = {arXiv},
  eprint       = {2310.01405},
  bibsource    = {dblp computer science bibliography, https://dblp.org}
}

@inproceedings{chen-etal-2025-sheetdesigner,
    title = "{S}heet{D}esigner: {MLLM}-Powered Spreadsheet Layout Generation with Rule-Based and Vision-Based Reflection",
    author = "Chen, Qin  and
      Ren, Yuanyi  and
      Ma, Xiaojun  and
      Liu, Mugeng  and
      Han, Shi  and
      Zhang, Dongmei",
    editor = "Christodoulopoulos, Christos  and
      Chakraborty, Tanmoy  and
      Rose, Carolyn  and
      Peng, Violet",
    booktitle = "Proceedings of the 2025 Conference on Empirical Methods in Natural Language Processing",
    month = nov,
    year = "2025",
    address = "Suzhou, China",
    publisher = "Association for Computational Linguistics",
    url = "https://aclanthology.org/2025.emnlp-main.957/",
    doi = "10.18653/v1/2025.emnlp-main.957",
    pages = "18921--18939",
    ISBN = "979-8-89176-332-6"
}

@inproceedings{chen-etal-2025-large-language,
    title = "Large Language Models for Predictive Analysis: How Far Are They?",
    author = "Chen, Qin  and
      Ren, Yuanyi  and
      Ma, Xiaojun  and
      Shi, Yuyang",
    editor = "Che, Wanxiang  and
      Nabende, Joyce  and
      Shutova, Ekaterina  and
      Pilehvar, Mohammad Taher",
    booktitle = "Findings of the Association for Computational Linguistics: ACL 2025",
    month = jul,
    year = "2025",
    address = "Vienna, Austria",
    publisher = "Association for Computational Linguistics",
    url = "https://aclanthology.org/2025.findings-acl.416/",
    doi = "10.18653/v1/2025.findings-acl.416",
    pages = "7961--7978",
    ISBN = "979-8-89176-256-5"
}

@article{DBLP:journals/corr/abs-2506-05346,
  author       = {Lei Hsiung and
                  Tianyu Pang and
                  Yung{-}Chen Tang and
                  Linyue Song and
                  Tsung{-}Yi Ho and
                  Pin{-}Yu Chen and
                  Yaoqing Yang},
  title        = {Why {LLM} Safety Guardrails Collapse After Fine-tuning: {A} Similarity
                  Analysis Between Alignment and Fine-tuning Datasets},
  journal      = {CoRR},
  volume       = {abs/2506.05346},
  year         = {2025},
  url          = {https://doi.org/10.48550/arXiv.2506.05346},
  doi          = {10.48550/ARXIV.2506.05346},
  eprinttype   = {arXiv},
  eprint       = {2506.05346},
  bibsource    = {dblp computer science bibliography, https://dblp.org}
}

@inproceedings{madaan2022emnlp,
  author       = {Aman Madaan and
                  Shuyan Zhou and
                  Uri Alon and
                  Yiming Yang and
                  Graham Neubig},
  editor       = {Yoav Goldberg and
                  Zornitsa Kozareva and
                  Yue Zhang},
  title        = {Language Models of Code are Few-Shot Commonsense Learners},
  booktitle    = {Proceedings of the 2022 Conference on Empirical Methods in Natural
                  Language Processing, {EMNLP} 2022, Abu Dhabi, United Arab Emirates,
                  December 7-11, 2022},
  pages        = {1384--1403},
  publisher    = {Association for Computational Linguistics},
  year         = {2022},
  url          = {https://doi.org/10.18653/v1/2022.emnlp-main.90},
  doi          = {10.18653/V1/2022.EMNLP-MAIN.90},
  bibsource    = {dblp computer science bibliography, https://dblp.org}
}

@inproceedings{
yu2024rankrag,
title={Rank{RAG}: Unifying Context Ranking with Retrieval-Augmented Generation in {LLM}s},
author={Yue Yu and Wei Ping and Zihan Liu and Boxin Wang and Jiaxuan You and Chao Zhang and Mohammad Shoeybi and Bryan Catanzaro},
booktitle={The Thirty-eighth Annual Conference on Neural Information Processing Systems},
year={2024},
url={https://openreview.net/forum?id=S1fc92uemC}
}

@inproceedings{DBLP:conf/conll/LevySCZ17,
  author       = {Omer Levy and
                  Minjoon Seo and
                  Eunsol Choi and
                  Luke Zettlemoyer},
  editor       = {Roger Levy and
                  Lucia Specia},
  title        = {Zero-Shot Relation Extraction via Reading Comprehension},
  booktitle    = {Proceedings of the 21st Conference on Computational Natural Language
                  Learning (CoNLL 2017), Vancouver, Canada, August 3-4, 2017},
  pages        = {333--342},
  publisher    = {Association for Computational Linguistics},
  year         = {2017},
  url          = {https://doi.org/10.18653/v1/K17-1034},
  doi          = {10.18653/V1/K17-1034},
  bibsource    = {dblp computer science bibliography, https://dblp.org}
}

@inproceedings{rimsky-etal-2024-steering,
    title = "Steering Llama 2 via Contrastive Activation Addition",
    author = "Rimsky, Nina  and
      Gabrieli, Nick  and
      Schulz, Julian  and
      Tong, Meg  and
      Hubinger, Evan  and
      Turner, Alexander",
    editor = "Ku, Lun-Wei  and
      Martins, Andre  and
      Srikumar, Vivek",
    booktitle = "Proceedings of the 62nd Annual Meeting of the Association for Computational Linguistics (Volume 1: Long Papers)",
    month = aug,
    year = "2024",
    address = "Bangkok, Thailand",
    publisher = "Association for Computational Linguistics",
    url = "https://aclanthology.org/2024.acl-long.828/",
    doi = "10.18653/v1/2024.acl-long.828",
    pages = "15504--15522"
}

@misc{
marks2024the,
title={The Geometry of Truth: Emergent Linear Structure in Large Language Model Representations of True/False Datasets},
author={Samuel Marks and Max Tegmark},
year={2024},
url={https://openreview.net/forum?id=CeJEfNKstt}
}

@inproceedings{NEURIPS2023_95b6e2ff,
 author = {Hartvigsen, Tom and Sankaranarayanan, Swami and Palangi, Hamid and Kim, Yoon and Ghassemi, Marzyeh},
 booktitle = {Advances in Neural Information Processing Systems},
 editor = {A. Oh and T. Naumann and A. Globerson and K. Saenko and M. Hardt and S. Levine},
 pages = {47934--47959},
 publisher = {Curran Associates, Inc.},
 title = {Aging with GRACE: Lifelong Model Editing with Discrete Key-Value Adaptors},
 url = {https://proceedings.neurips.cc/paper_files/paper/2023/file/95b6e2ff961580e03c0a662a63a71812-Paper-Conference.pdf},
 volume = {36},
 year = {2023}
}

@inproceedings{
li2025taming,
title={Taming Knowledge Conflicts in Language Models},
author={Gaotang Li and Yuzhong Chen and Hanghang Tong},
booktitle={Forty-second International Conference on Machine Learning},
year={2025},
url={https://openreview.net/forum?id=0cEZyhHEks}
}

@article{gutierrez2025rag,
  title={From rag to memory: Non-parametric continual learning for large language models},
  author={Guti{\'e}rrez, Bernal Jim{\'e}nez and Shu, Yiheng and Qi, Weijian and Zhou, Sizhe and Su, Yu},
  journal={arXiv preprint arXiv:2502.14802},
  year={2025}
}

@inproceedings{sun-etal-2024-head,
    title = "Head-to-Tail: How Knowledgeable are Large Language Models ({LLM}s)? {A}.{K}.{A}. Will {LLM}s Replace Knowledge Graphs?",
    author = "Sun, Kai  and
      Xu, Yifan  and
      Zha, Hanwen  and
      Liu, Yue  and
      Dong, Xin Luna",
    editor = "Duh, Kevin  and
      Gomez, Helena  and
      Bethard, Steven",
    booktitle = "Proceedings of the 2024 Conference of the North American Chapter of the Association for Computational Linguistics: Human Language Technologies (Volume 1: Long Papers)",
    month = jun,
    year = "2024",
    address = "Mexico City, Mexico",
    publisher = "Association for Computational Linguistics",
    url = "https://aclanthology.org/2024.naacl-long.18/",
    doi = "10.18653/v1/2024.naacl-long.18",
    pages = "311--325"
}

@misc{
nishi2025representation,
title={Representation Shattering in Transformers: A Synthetic Study with Knowledge Editing},
author={Kento Nishi and Maya Okawa and Rahul Ramesh and Mikail Khona and Hidenori Tanaka and Ekdeep Singh Lubana},
year={2025},
url={https://openreview.net/forum?id=MjFoQAhnl3}
}

@inproceedings{yang2024acl,
  author       = {Wanli Yang and
                  Fei Sun and
                  Xinyu Ma and
                  Xun Liu and
                  Dawei Yin and
                  Xueqi Cheng},
  editor       = {Lun{-}Wei Ku and
                  Andre Martins and
                  Vivek Srikumar},
  title        = {The Butterfly Effect of Model Editing: Few Edits Can Trigger Large
                  Language Models Collapse},
  booktitle    = {Findings of the Association for Computational Linguistics, {ACL} 2024,
                  Bangkok, Thailand and virtual meeting, August 11-16, 2024},
  pages        = {5419--5437},
  publisher    = {Association for Computational Linguistics},
  year         = {2024},
  url          = {https://doi.org/10.18653/v1/2024.findings-acl.322},
  doi          = {10.18653/V1/2024.FINDINGS-ACL.322},
  bibsource    = {dblp computer science bibliography, https://dblp.org}
}

@inproceedings{ke2023iclr,
  author       = {Zixuan Ke and
                  Yijia Shao and
                  Haowei Lin and
                  Tatsuya Konishi and
                  Gyuhak Kim and
                  Bing Liu},
  title        = {Continual Pre-training of Language Models},
  booktitle    = {The Eleventh International Conference on Learning Representations,
                  {ICLR} 2023, Kigali, Rwanda, May 1-5, 2023},
  publisher    = {OpenReview.net},
  year         = {2023},
  url          = {https://openreview.net/forum?id=m\_GDIItaI3o},
  bibsource    = {dblp computer science bibliography, https://dblp.org}
}

@article{yildiz2025tmlr,
  author       = {{\c{C}}agatay Yildiz and
                  Nishaanth Kanna Ravichandran and
                  Nitin Sharma and
                  Matthias Bethge and
                  Beyza Ermis},
  title        = {Investigating Continual Pretraining in Large Language Models: Insights
                  and Implications},
  journal      = {Trans. Mach. Learn. Res.},
  volume       = {2025},
  year         = {2025},
  url          = {https://openreview.net/forum?id=aKjJoEVKgO},
  bibsource    = {dblp computer science bibliography, https://dblp.org}
}

@inproceedings{salemiza2024sigir,
  author       = {Alireza Salemi and
                  Hamed Zamani},
  editor       = {Grace Hui Yang and
                  Hongning Wang and
                  Sam Han and
                  Claudia Hauff and
                  Guido Zuccon and
                  Yi Zhang},
  title        = {Evaluating Retrieval Quality in Retrieval-Augmented Generation},
  booktitle    = {Proceedings of the 47th International {ACM} {SIGIR} Conference on
                  Research and Development in Information Retrieval, {SIGIR} 2024, Washington
                  DC, USA, July 14-18, 2024},
  pages        = {2395--2400},
  publisher    = {{ACM}},
  year         = {2024},
  url          = {https://doi.org/10.1145/3626772.3657957},
  doi          = {10.1145/3626772.3657957},
  bibsource    = {dblp computer science bibliography, https://dblp.org}
}

@inproceedings{DBLP:conf/iclr/HendrycksBBZMSS21,
  author       = {Dan Hendrycks and
                  Collin Burns and
                  Steven Basart and
                  Andy Zou and
                  Mantas Mazeika and
                  Dawn Song and
                  Jacob Steinhardt},
  title        = {Measuring Massive Multitask Language Understanding},
  booktitle    = {9th International Conference on Learning Representations, {ICLR} 2021,
                  Virtual Event, Austria, May 3-7, 2021},
  publisher    = {OpenReview.net},
  year         = {2021},
  url          = {https://openreview.net/forum?id=d7KBjmI3GmQ},
  bibsource    = {dblp computer science bibliography, https://dblp.org}
}

@article{turner2023steering,
  title={Steering language models with activation engineering},
  author={Turner, Alexander Matt and Thiergart, Lisa and Leech, Gavin and Udell, David and Vazquez, Juan J and Mini, Ulisse and MacDiarmid, Monte},
  journal={arXiv preprint arXiv:2308.10248},
  year={2023}
}

@article{DBLP:journals/corr/abs-2310-06824,
  author       = {Samuel Marks and
                  Max Tegmark},
  title        = {The Geometry of Truth: Emergent Linear Structure in Large Language
                  Model Representations of True/False Datasets},
  journal      = {CoRR},
  volume       = {abs/2310.06824},
  year         = {2023},
  url          = {https://doi.org/10.48550/arXiv.2310.06824},
  doi          = {10.48550/ARXIV.2310.06824},
  eprinttype    = {arXiv},
  eprint       = {2310.06824},
  bibsource    = {dblp computer science bibliography, https://dblp.org}
}

@inproceedings{DBLP:conf/nips/ArditiOSPPGN24,
  author       = {Andy Arditi and
                  Oscar Obeso and
                  Aaquib Syed and
                  Daniel Paleka and
                  Nina Panickssery and
                  Wes Gurnee and
                  Neel Nanda},
  editor       = {Amir Globersons and
                  Lester Mackey and
                  Danielle Belgrave and
                  Angela Fan and
                  Ulrich Paquet and
                  Jakub M. Tomczak and
                  Cheng Zhang},
  title        = {Refusal in Language Models Is Mediated by a Single Direction},
  booktitle    = {Advances in Neural Information Processing Systems 38: Annual Conference
                  on Neural Information Processing Systems 2024, NeurIPS 2024, Vancouver,
                  BC, Canada, December 10 - 15, 2024},
  year         = {2024},
  url          = {http://papers.nips.cc/paper\_files/paper/2024/hash/f545448535dfde4f9786555403ab7c49-Abstract-Conference.html},
  bibsource    = {dblp computer science bibliography, https://dblp.org}
}

@article{chen2025seal,
  title={Seal: Steerable reasoning calibration of large language models for free},
  author={Chen, Runjin and Zhang, Zhenyu and Hong, Junyuan and Kundu, Souvik and Wang, Zhangyang},
  journal={arXiv preprint arXiv:2504.07986},
  year={2025}
}

@inproceedings{DBLP:conf/acl/HanXL0SJAJ24,
  author       = {Chi Han and
                  Jialiang Xu and
                  Manling Li and
                  Yi Fung and
                  Chenkai Sun and
                  Nan Jiang and
                  Tarek F. Abdelzaher and
                  Heng Ji},
  editor       = {Lun{-}Wei Ku and
                  Andre Martins and
                  Vivek Srikumar},
  title        = {Word Embeddings Are Steers for Language Models},
  booktitle    = {Proceedings of the 62nd Annual Meeting of the Association for Computational
                  Linguistics (Volume 1: Long Papers), {ACL} 2024, Bangkok, Thailand,
                  August 11-16, 2024},
  pages        = {16410--16430},
  publisher    = {Association for Computational Linguistics},
  year         = {2024},
  url          = {https://doi.org/10.18653/v1/2024.acl-long.864},
  doi          = {10.18653/V1/2024.ACL-LONG.864},
  bibsource    = {dblp computer science bibliography, https://dblp.org}
}

@inproceedings{10.5555/3692070.3693675,
author = {Park, Kiho and Choe, Yo Joong and Veitch, Victor},
title = {The linear representation hypothesis and the geometry of large language models},
year = {2024},
publisher = {JMLR.org},
booktitle = {Proceedings of the 41st International Conference on Machine Learning},
articleno = {1605},
numpages = {24},
location = {Vienna, Austria},
series = {ICML'24}
}

@inproceedings{DBLP:conf/nips/0104L0XY0X0C24,
  author       = {Peng Wang and
                  Zexi Li and
                  Ningyu Zhang and
                  Ziwen Xu and
                  Yunzhi Yao and
                  Yong Jiang and
                  Pengjun Xie and
                  Fei Huang and
                  Huajun Chen},
  editor       = {Amir Globersons and
                  Lester Mackey and
                  Danielle Belgrave and
                  Angela Fan and
                  Ulrich Paquet and
                  Jakub M. Tomczak and
                  Cheng Zhang},
  title        = {{WISE:} Rethinking the Knowledge Memory for Lifelong Model Editing
                  of Large Language Models},
  booktitle    = {Advances in Neural Information Processing Systems 38: Annual Conference
                  on Neural Information Processing Systems 2024, NeurIPS 2024, Vancouver,
                  BC, Canada, December 10 - 15, 2024},
  year         = {2024},
  url          = {http://papers.nips.cc/paper\_files/paper/2024/hash/60960ad78868fce5c165295fbd895060-Abstract-Conference.html},
  bibsource    = {dblp computer science bibliography, https://dblp.org}
}

@inproceedings{
liu2025unlocking,
title={Unlocking Efficient, Scalable, and Continual Knowledge Editing with Basis-Level Representation Fine-Tuning},
author={Tianci Liu and Ruirui Li and Yunzhe Qi and Hui Liu and Xianfeng Tang and Tianqi Zheng and Qingyu Yin and Monica Xiao Cheng and Jun Huan and Haoyu Wang and Jing Gao},
booktitle={The Thirteenth International Conference on Learning Representations},
year={2025},
url={https://openreview.net/forum?id=PITFO1ddeh}
}

@article{DBLP:journals/corr/abs-2502-05628,
  author       = {Houcheng Jiang and
                  Junfeng Fang and
                  Ningyu Zhang and
                  Guojun Ma and
                  Mingyang Wan and
                  Xiang Wang and
                  Xiangnan He and
                  Tat{-}Seng Chua},
  title        = {AnyEdit: Edit Any Knowledge Encoded in Language Models},
  journal      = {CoRR},
  volume       = {abs/2502.05628},
  year         = {2025},
  url          = {https://doi.org/10.48550/arXiv.2502.05628},
  doi          = {10.48550/ARXIV.2502.05628},
  eprinttype    = {arXiv},
  eprint       = {2502.05628},
  bibsource    = {dblp computer science bibliography, https://dblp.org}
}

@inproceedings{mikolov-etal-2013-linguistic,
    title = "Linguistic Regularities in Continuous Space Word Representations",
    author = "Mikolov, Tomas  and
      Yih, Wen-tau  and
      Zweig, Geoffrey",
    editor = "Vanderwende, Lucy  and
      Daum{\'e} III, Hal  and
      Kirchhoff, Katrin",
    booktitle = "Proceedings of the 2013 Conference of the North {A}merican Chapter of the Association for Computational Linguistics: Human Language Technologies",
    month = jun,
    year = "2013",
    address = "Atlanta, Georgia",
    publisher = "Association for Computational Linguistics",
    url = "https://aclanthology.org/N13-1090/",
    pages = "746--751"
}

@article{elhage2021mathematical,
   title={A Mathematical Framework for Transformer Circuits},
   author={Elhage, Nelson and Nanda, Neel and Olsson, Catherine and Henighan, Tom and Joseph, Nicholas and Mann, Ben and Askell, Amanda and Bai, Yuntao and Chen, Anna and Conerly, Tom and DasSarma, Nova and Drain, Dawn and Ganguli, Deep and Hatfield-Dodds, Zac and Hernandez, Danny and Jones, Andy and Kernion, Jackson and Lovitt, Liane and Ndousse, Kamal and Amodei, Dario and Brown, Tom and Clark, Jack and Kaplan, Jared and McCandlish, Sam and Olah, Chris},
   year={2021},
   journal={Transformer Circuits Thread},
   note={https://transformer-circuits.pub/2021/framework/index.html}
}

@article{zhu2020modifying,
  title={Modifying memories in transformer models},
  author={Zhu, Chen and Rawat, Ankit Singh and Zaheer, Manzil and Bhojanapalli, Srinadh and Li, Daliang and Yu, Felix and Kumar, Sanjiv},
  journal={arXiv preprint arXiv:2012.00363},
  year={2020}
}

@inproceedings{cheng-etal-2025-serial,
    title = "Serial Lifelong Editing via Mixture of Knowledge Experts",
    author = "Cheng, YuJu  and
      Yu, Yu-Chu  and
      Chang, Kai-Po  and
      Wang, Yu-Chiang Frank",
    editor = "Che, Wanxiang  and
      Nabende, Joyce  and
      Shutova, Ekaterina  and
      Pilehvar, Mohammad Taher",
    booktitle = "Proceedings of the 63rd Annual Meeting of the Association for Computational Linguistics (Volume 1: Long Papers)",
    month = jul,
    year = "2025",
    address = "Vienna, Austria",
    publisher = "Association for Computational Linguistics",
    url = "https://aclanthology.org/2025.acl-long.1492/",
    doi = "10.18653/v1/2025.acl-long.1492",
    pages = "30888--30903",
    ISBN = "979-8-89176-251-0"
}

@inproceedings{
li2024badedit,
title={BadEdit: Backdooring Large Language Models by Model Editing},
author={Yanzhou Li and Tianlin Li and Kangjie Chen and Jian Zhang and Shangqing Liu and Wenhan Wang and Tianwei Zhang and Yang Liu},
booktitle={The Twelfth International Conference on Learning Representations},
year={2024},
url={https://openreview.net/forum?id=duZANm2ABX}
}

@article{cohen2024tacl,
  author       = {Roi Cohen and
                  Eden Biran and
                  Ori Yoran and
                  Amir Globerson and
                  Mor Geva},
  title        = {Evaluating the Ripple Effects of Knowledge Editing in Language Models},
  journal      = {Trans. Assoc. Comput. Linguistics},
  volume       = {12},
  pages        = {283--298},
  year         = {2024},
  url          = {https://doi.org/10.1162/tacl\_a\_00644},
  doi          = {10.1162/TACL\_A\_00644},
  bibsource    = {dblp computer science bibliography, https://dblp.org}
}

@inproceedings{liu-etal-2025-spectral,
    title = "Spectral Insights into Data-Oblivious Critical Layers in Large Language Models",
    author = "Liu, Xuyuan  and
      Hsiung, Lei  and
      Yang, Yaoqing  and
      Yan, Yujun",
    editor = "Che, Wanxiang  and
      Nabende, Joyce  and
      Shutova, Ekaterina  and
      Pilehvar, Mohammad Taher",
    booktitle = "Findings of the Association for Computational Linguistics: ACL 2025",
    month = jul,
    year = "2025",
    address = "Vienna, Austria",
    publisher = "Association for Computational Linguistics",
    url = "https://aclanthology.org/2025.findings-acl.251/",
    doi = "10.18653/v1/2025.findings-acl.251",
    pages = "4860--4877",
    ISBN = "979-8-89176-256-5"
}

@inproceedings{chen-etal-2024-lifelong,
    title = "Lifelong Knowledge Editing for {LLM}s with Retrieval-Augmented Continuous Prompt Learning",
    author = "Chen, Qizhou  and
      Zhang, Taolin  and
      He, Xiaofeng  and
      Li, Dongyang  and
      Wang, Chengyu  and
      Huang, Longtao  and
      Xue{'}, Hui",
    editor = "Al-Onaizan, Yaser  and
      Bansal, Mohit  and
      Chen, Yun-Nung",
    booktitle = "Proceedings of the 2024 Conference on Empirical Methods in Natural Language Processing",
    month = nov,
    year = "2024",
    address = "Miami, Florida, USA",
    publisher = "Association for Computational Linguistics",
    url = "https://aclanthology.org/2024.emnlp-main.751/",
    doi = "10.18653/v1/2024.emnlp-main.751",
    pages = "13565--13580"
}

@inproceedings{
thede2025wikibigedit,
title={WikiBigEdit: Understanding the Limits of Lifelong Knowledge Editing in {LLM}s},
author={Lukas Thede and Karsten Roth and Matthias Bethge and Zeynep Akata and Thomas Hartvigsen},
booktitle={Forty-second International Conference on Machine Learning},
year={2025},
url={https://openreview.net/forum?id=9NVm1Bf7CS}
}

@inproceedings{geva-etal-2021-transformer,
    title = "Transformer Feed-Forward Layers Are Key-Value Memories",
    author = "Geva, Mor  and
      Schuster, Roei  and
      Berant, Jonathan  and
      Levy, Omer",
    editor = "Moens, Marie-Francine  and
      Huang, Xuanjing  and
      Specia, Lucia  and
      Yih, Scott Wen-tau",
    booktitle = "Proceedings of the 2021 Conference on Empirical Methods in Natural Language Processing",
    month = nov,
    year = "2021",
    address = "Online and Punta Cana, Dominican Republic",
    publisher = "Association for Computational Linguistics",
    url = "https://aclanthology.org/2021.emnlp-main.446/",
    doi = "10.18653/v1/2021.emnlp-main.446",
    pages = "5484--5495"
}

@article{Yang2024Qwen25TR,
  title={Qwen2.5 Technical Report},
  author={Qwen An Yang and Baosong Yang and Beichen Zhang and Binyuan Hui and Bo Zheng and Bowen Yu and Chengyuan Li and Dayiheng Liu and Fei Huang and Guanting Dong and Haoran Wei and Huan Lin and Jian Yang and Jianhong Tu and Jianwei Zhang and Jianxin Yang and Jiaxin Yang and Jingren Zhou and Junyang Lin and Kai Dang and Keming Lu and Keqin Bao and Kexin Yang and Le Yu and Mei Li and Mingfeng Xue and Pei Zhang and Qin Zhu and Rui Men and Runji Lin and Tianhao Li and Tingyu Xia and Xingzhang Ren and Xuancheng Ren and Yang Fan and Yang Su and Yi-Chao Zhang and Yunyang Wan and Yuqi Liu and Zeyu Cui and Zhenru Zhang and Zihan Qiu and Shanghaoran Quan and Zekun Wang},
  journal={ArXiv},
  year={2024},
  volume={abs/2412.15115},
  url={https://api.semanticscholar.org/CorpusID:274859421}
}

@article{grattafiori2024llama,
  title={The llama 3 herd of models},
  author={Grattafiori, Aaron and Dubey, Abhimanyu and Jauhri, Abhinav and Pandey, Abhinav and Kadian, Abhishek and Al-Dahle, Ahmad and Letman, Aiesha and Mathur, Akhil and Schelten, Alan and Vaughan, Alex and others},
  journal={arXiv preprint arXiv:2407.21783},
  year={2024}
}

@inproceedings{DBLP:conf/nips/WuAWGJMP24,
  author       = {Zhengxuan Wu and
                  Aryaman Arora and
                  Zheng Wang and
                  Atticus Geiger and
                  Dan Jurafsky and
                  Christopher D. Manning and
                  Christopher Potts},
  editor       = {Amir Globersons and
                  Lester Mackey and
                  Danielle Belgrave and
                  Angela Fan and
                  Ulrich Paquet and
                  Jakub M. Tomczak and
                  Cheng Zhang},
  title        = {ReFT: Representation Finetuning for Language Models},
  booktitle    = {Advances in Neural Information Processing Systems 38: Annual Conference
                  on Neural Information Processing Systems 2024, NeurIPS 2024, Vancouver,
                  BC, Canada, December 10 - 15, 2024},
  year         = {2024},
  url          = {http://papers.nips.cc/paper\_files/paper/2024/hash/75008a0fba53bf13b0bb3b7bff986e0e-Abstract-Conference.html},
  bibsource    = {dblp computer science bibliography, https://dblp.org}
}

@misc{chen2025personavectorsmonitoringcontrolling,
      title={Persona Vectors: Monitoring and Controlling Character Traits in Language Models}, 
      author={Runjin Chen and Andy Arditi and Henry Sleight and Owain Evans and Jack Lindsey},
      year={2025},
      eprint={2507.21509},
      archivePrefix={arXiv},
      primaryClass={cs.CL},
      url={https://arxiv.org/abs/2507.21509}, 
}

@inproceedings{DBLP:conf/iclr/MitchellLBFM22,
  author       = {Eric Mitchell and
                  Charles Lin and
                  Antoine Bosselut and
                  Chelsea Finn and
                  Christopher D. Manning},
  title        = {Fast Model Editing at Scale},
  booktitle    = {The Tenth International Conference on Learning Representations, {ICLR}
                  2022, Virtual Event, April 25-29, 2022},
  publisher    = {OpenReview.net},
  year         = {2022},
  url          = {https://openreview.net/forum?id=0DcZxeWfOPt},
  bibsource    = {dblp computer science bibliography, https://dblp.org}
}

@inproceedings{DBLP:conf/emnlp/ZhengLDFWXC23,
  author       = {Ce Zheng and
                  Lei Li and
                  Qingxiu Dong and
                  Yuxuan Fan and
                  Zhiyong Wu and
                  Jingjing Xu and
                  Baobao Chang},
  editor       = {Houda Bouamor and
                  Juan Pino and
                  Kalika Bali},
  title        = {Can We Edit Factual Knowledge by In-Context Learning?},
  booktitle    = {Proceedings of the 2023 Conference on Empirical Methods in Natural
                  Language Processing, {EMNLP} 2023, Singapore, December 6-10, 2023},
  pages        = {4862--4876},
  publisher    = {Association for Computational Linguistics},
  year         = {2023},
  url          = {https://doi.org/10.18653/v1/2023.emnlp-main.296},
  doi          = {10.18653/V1/2023.EMNLP-MAIN.296},
  bibsource    = {dblp computer science bibliography, https://dblp.org}
}

@inproceedings{DBLP:conf/nips/MengBAB22,
  author       = {Kevin Meng and
                  David Bau and
                  Alex Andonian and
                  Yonatan Belinkov},
  editor       = {Sanmi Koyejo and
                  S. Mohamed and
                  A. Agarwal and
                  Danielle Belgrave and
                  K. Cho and
                  A. Oh},
  title        = {Locating and Editing Factual Associations in {GPT}},
  booktitle    = {Advances in Neural Information Processing Systems 35: Annual Conference
                  on Neural Information Processing Systems 2022, NeurIPS 2022, New Orleans,
                  LA, USA, November 28 - December 9, 2022},
  year         = {2022},
  url          = {http://papers.nips.cc/paper\_files/paper/2022/hash/6f1d43d5a82a37e89b0665b33bf3a182-Abstract-Conference.html},
  bibsource    = {dblp computer science bibliography, https://dblp.org}
}

@article{DBLP:journals/corr/abs-2210-07229,
  author       = {Kevin Meng and
                  Arnab Sen Sharma and
                  Alex Andonian and
                  Yonatan Belinkov and
                  David Bau},
  title        = {Mass-Editing Memory in a Transformer},
  journal      = {CoRR},
  volume       = {abs/2210.07229},
  year         = {2022},
  url          = {https://doi.org/10.48550/arXiv.2210.07229},
  doi          = {10.48550/ARXIV.2210.07229},
  eprinttype    = {arXiv},
  eprint       = {2210.07229},
  bibsource    = {dblp computer science bibliography, https://dblp.org}
}

@inproceedings{DBLP:conf/iclr/FangJWMSW0C25,
  author       = {Junfeng Fang and
                  Houcheng Jiang and
                  Kun Wang and
                  Yunshan Ma and
                  Jie Shi and
                  Xiang Wang and
                  Xiangnan He and
                  Tat{-}Seng Chua},
  title        = {AlphaEdit: Null-Space Constrained Knowledge Editing for Language Models},
  booktitle    = {The Thirteenth International Conference on Learning Representations,
                  {ICLR} 2025, Singapore, April 24-28, 2025},
  publisher    = {OpenReview.net},
  year         = {2025},
  url          = {https://openreview.net/forum?id=HvSytvg3Jh},
  bibsource    = {dblp computer science bibliography, https://dblp.org}
}

@inproceedings{DBLP:conf/nips/HartvigsenSPKG23,
  author       = {Tom Hartvigsen and
                  Swami Sankaranarayanan and
                  Hamid Palangi and
                  Yoon Kim and
                  Marzyeh Ghassemi},
  editor       = {Alice Oh and
                  Tristan Naumann and
                  Amir Globerson and
                  Kate Saenko and
                  Moritz Hardt and
                  Sergey Levine},
  title        = {Aging with {GRACE:} Lifelong Model Editing with Discrete Key-Value
                  Adaptors},
  booktitle    = {Advances in Neural Information Processing Systems 36: Annual Conference
                  on Neural Information Processing Systems 2023, NeurIPS 2023, New Orleans,
                  LA, USA, December 10 - 16, 2023},
  year         = {2023},
  url          = {http://papers.nips.cc/paper\_files/paper/2023/hash/95b6e2ff961580e03c0a662a63a71812-Abstract-Conference.html},
  bibsource    = {dblp computer science bibliography, https://dblp.org}
}

@article{DBLP:journals/corr/abs-2502-00158,
  author       = {Binchi Zhang and
                  Zhengzhang Chen and
                  Zaiyi Zheng and
                  Jundong Li and
                  Haifeng Chen},
  title        = {Resolving Editing-Unlearning Conflicts: {A} Knowledge Codebook Framework
                  for Large Language Model Updating},
  journal      = {CoRR},
  volume       = {abs/2502.00158},
  year         = {2025},
  url          = {https://doi.org/10.48550/arXiv.2502.00158},
  doi          = {10.48550/ARXIV.2502.00158},
  eprinttype    = {arXiv},
  eprint       = {2502.00158},
  bibsource    = {dblp computer science bibliography, https://dblp.org}
}

@inproceedings{DBLP:conf/iclr/DengWPDSC25,
  author       = {Jingcheng Deng and
                  Zihao Wei and
                  Liang Pang and
                  Hanxing Ding and
                  Huawei Shen and
                  Xueqi Cheng},
  title        = {Everything is Editable: Extend Knowledge Editing to Unstructured Data
                  in Large Language Models},
  booktitle    = {The Thirteenth International Conference on Learning Representations,
                  {ICLR} 2025, Singapore, April 24-28, 2025},
  publisher    = {OpenReview.net},
  year         = {2025},
  url          = {https://openreview.net/forum?id=X5rO5VyTgB},
  bibsource    = {dblp computer science bibliography, https://dblp.org}
}

@inproceedings{DBLP:conf/emnlp/WuPWL24,
  author       = {Xiaobao Wu and
                  Liangming Pan and
                  William Yang Wang and
                  Anh Tuan Luu},
  editor       = {Yaser Al{-}Onaizan and
                  Mohit Bansal and
                  Yun{-}Nung Chen},
  title        = {{AKEW:} Assessing Knowledge Editing in the Wild},
  booktitle    = {Proceedings of the 2024 Conference on Empirical Methods in Natural
                  Language Processing, {EMNLP} 2024, Miami, FL, USA, November 12-16,
                  2024},
  pages        = {15118--15133},
  publisher    = {Association for Computational Linguistics},
  year         = {2024},
  url          = {https://doi.org/10.18653/v1/2024.emnlp-main.843},
  doi          = {10.18653/V1/2024.EMNLP-MAIN.843},
  bibsource    = {dblp computer science bibliography, https://dblp.org}
}

@article{DBLP:journals/corr/abs-2506-07899,
  author       = {Ke Wang and
                  Yiming Qin and
                  Nikolaos Dimitriadis and
                  Alessandro Favero and
                  Pascal Frossard},
  title        = {{MEMOIR:} Lifelong Model Editing with Minimal Overwrite and Informed
                  Retention for LLMs},
  journal      = {CoRR},
  volume       = {abs/2506.07899},
  year         = {2025},
  url          = {https://doi.org/10.48550/arXiv.2506.07899},
  doi          = {10.48550/ARXIV.2506.07899},
  eprinttype    = {arXiv},
  eprint       = {2506.07899},
  bibsource    = {dblp computer science bibliography, https://dblp.org}
}
